\documentclass{article}
\usepackage[utf8]{inputenc}
\usepackage[T1]{fontenc}
\usepackage{main}
\usepackage{microtype}
\usepackage{graphicx}
\usepackage{times}
\usepackage{amsmath}
\usepackage{enumitem}
\usepackage{booktabs}
\usepackage{xcolor}     
\usepackage{natbib}
\definecolor{mydarkblue}{rgb}{0,0.08,0.45}
\usepackage[colorlinks=true,linkcolor=mydarkblue,citecolor=mydarkblue,filecolor=mydarkblue,urlcolor=mydarkblue]{hyperref}
\usepackage{fancyhdr}
\usepackage{amsmath}
\usepackage{bbm}
\usepackage{wrapfig}
\usepackage{graphicx}
\usepackage{multirow} 
\usepackage{makecell}
\usepackage{fvextra}            
\usepackage[table, dvipsnames]{xcolor}
\usepackage{booktabs}
\usepackage{alltt}
\usepackage{xspace}
\usepackage{caption}
\usepackage{adjustbox}
\usepackage{verbatim}
\usepackage[most]{tcolorbox}
\tcbuselibrary{listings,breakable}

\usepackage{longtable}
\usepackage{siunitx}
\RecustomVerbatimEnvironment{verbatim}{Verbatim}{
  breaklines,       
  breakanywhere,    
  formatcom=\ttfamily\small 
}
\usepackage{booktabs}
\usepackage{wrapfig}
\usepackage{tabularx}

\usepackage{amssymb}
\usepackage{makecell}
\renewcommand{\arraystretch}{1.19}
\definecolor{instancepink}{RGB}{255, 220, 220}
\definecolor{modelgray}{RGB}{245,245,245}
\usepackage{caption}
\usepackage{tcolorbox}
\usepackage{listings}
\definecolor{instancepink}{RGB}{255, 220, 220}
\definecolor{modelbox}{RGB}{245,245,245}
\definecolor{myred}{RGB}{180,0,0}

\usepackage{listings}
\usepackage{xcolor}

\lstdefinelanguage{json}{
    basicstyle=\ttfamily\footnotesize,
    breaklines=true,
    breakatwhitespace=true,
    columns=fullflexible,
    keepspaces=true,
    showstringspaces=false
}

\newcommand{\ourmethod}{\textsc{CAID}\xspace}

\DefineVerbatimEnvironment{code}{Verbatim}{
  breaklines,                   
  breakanywhere=true,   
  breaksymbolleft={},  
  breaksymbolright={},
  breakindent=0pt,
}

\usepackage{color-edits}
\addauthor{gn}{magenta}

\definecolor{gred}{RGB}{250, 210, 207}
\definecolor{coolblue1}{rgb}{0.91, 0.94, 0.98}
\definecolor{coolblue2}{rgb}{0.76, 0.85, 0.94}
\definecolor{coolblue3}{rgb}{0.54, 0.72, 0.87}
\definecolor{coolblue4}{rgb}{1, 1, 1}

\usepackage{xspace}
\usepackage{cleveref}

\newenvironment{itemize*}%
 {\leftmargini=10pt\begin{itemize}%
  \setlength{\itemsep}{0pt}%
  \setlength{\parskip}{0pt}%
  }%
 {\end{itemize}}
\newenvironment{enumerate*}%
 {\begin{enumerate}%
  \setlength{\itemsep}{0pt}%
  \setlength{\parskip}{0pt}}%
 {\end{enumerate}}

\begin{document}

\title{\textbf{Effective Strategies for \\ Asynchronous Software Engineering Agents}}

\author{
\textbf{Jiayi Geng}$^{1}$ \quad
\textbf{Graham Neubig}$^{1}$ \quad \\
\textsuperscript{1}Carnegie Mellon University, Language Technologies Institute \\
\texttt{\{ogeng, gneubig\}@cs.cmu.edu} \\\\
    \href{https://github.com/JiayiGeng/async-swe-agents}{\includegraphics[height=0.4cm]{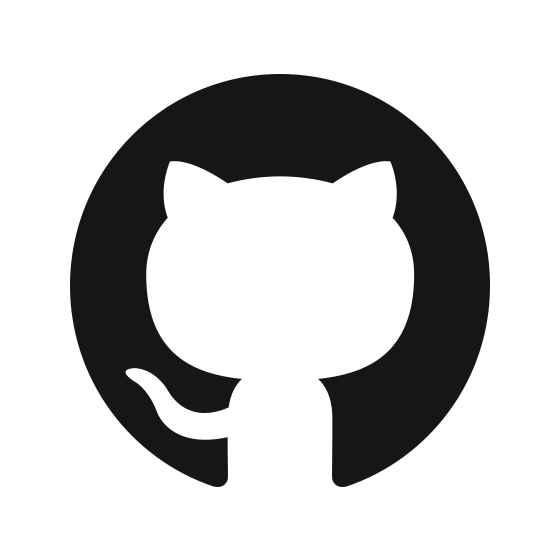} \textbf{https://github.com/JiayiGeng/CAID}} ~ ~ ~ 
}

\maketitle
\thispagestyle{fancy}
\fancyhead{}
\lhead{\includegraphics[height=0.5cm]{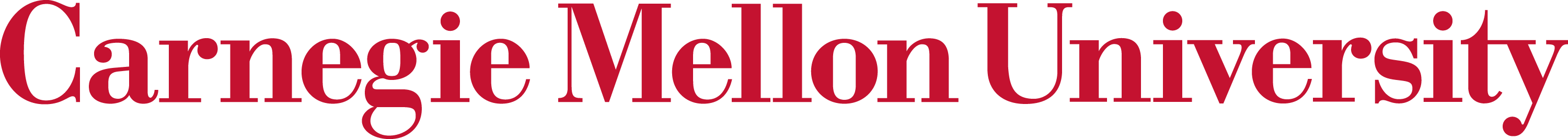}}
\rhead{%
  \raisebox{-0.1cm}{\includegraphics[height=0.8cm]{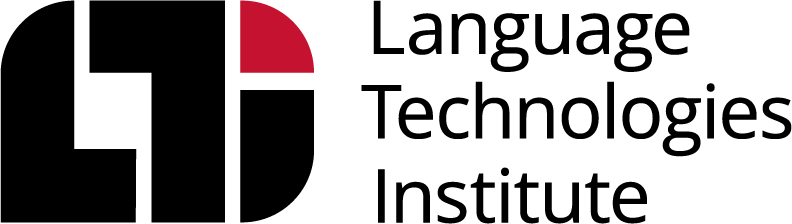}}%
}
\renewcommand{\headrulewidth}{0pt}
\setlength{\headheight}{12pt}
\addtolength{\topmargin}{00pt}
\setlength{\headsep}{3mm}

\vspace{-1.0em}
\begin{abstract}
AI agents have become increasingly capable at isolated software engineering (SWE) tasks such as resolving issues on Github.
Yet long-horizon tasks involving multiple interdependent subtasks still pose challenges both with respect to accuracy, and with respect to timely completion.
A natural approach to solving these long-horizon tasks in a timely manner is asynchronous multi-agent collaboration, where multiple agents work on different parts of the task at the same time.
But effective application of multi-agent systems has proven surprisingly difficult: concurrent edits by multiple agents interfere with each other, dependencies are difficult to synchronize, and combining partial progress into a coherent whole is challenging.
On the other hand, human developers have long relied on mature collaboration infrastructure to manage these challenges in large software projects.
Inspired by these collaboration primitives, we introduce \underline{\textit{C}}entralized \underline{\textit{A}}synchronous \underline{\textit{I}}solated \underline{\textit{D}}elegation (\ourmethod), a structured multi-agent coordination paradigm grounded in three core SWE primitives: centralized task delegation, asynchronous execution, and isolated workspaces. \ourmethod constructs dependency-aware task plans through a central manager, executes subtasks concurrently in isolated workspaces, and consolidates progress via structured integration with executable test-based verification. In empirical evaluation, we find that \ourmethod improves accuracy over single-agent baselines by 25.6\% absolute on paper reproduction tasks (PaperBench) and 14.7\% on Python library development tasks (Commit0).
Through systematic analysis, we find that branch-and-merge is a central coordination mechanism for multi-agent collaboration, and that SWE primitives such as \texttt{git worktree}, \texttt{git commit}, and \texttt{git merge} enable it to be realized in a reliable and executable manner.

\begin{figure}[h]
    \centering
    \includegraphics[width=0.95\textwidth]{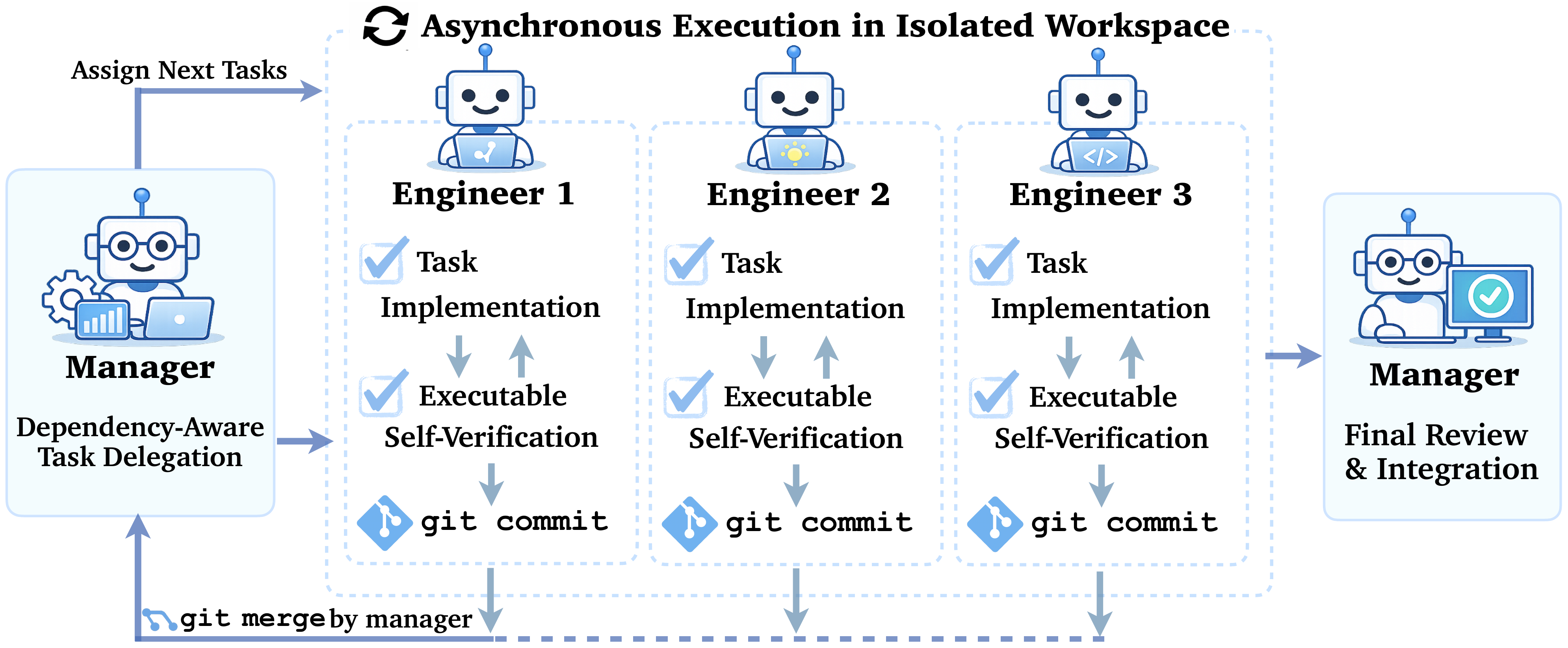}
    \caption{
       \textbf{Overview of \ourmethod Workflow.} The Manager explores the SWE tasks, builds a dependency graph to decompose tasks into parallelizable groups, and creates isolated git worktrees for every onboarded engineer. In the asynchronous loop, engineers independently implement, self-verify, and make a commit. Upon any engineer's completion, the Manager merges to main and dynamically updates the task delegation plan before reassigning the next task. After the asynchronous loop, the manager does a final review before submitting the final product. }
    \label{fig:teaser}
\end{figure}

\end{abstract}
\section{Introduction}
\label{sec:introduction}


As LLM-based software engineering agents improve, we have come to expect more of them.
Whereas fixing isolated github issues on real-world repositories was a major challenge a few years ago \citep{jimenez2023swe,yang2024swe,wang2024openhands}, we are now asking agents to build large apps from scratch \citep{zhao2024commit0} or implement entire research papers \citep{starace2025paperbench}. One method for performing this implementation is tasking a single agent with a large task, and hoping that it can execute on it from start to finish.
While task-completion horizons of agents continue to grow rapidly \citep{kwa2025measuring}, these systems are still limited in the scope of tasks they can perform reliably, and a single agent performing a large task also takes significant wall-clock time.
To this end, in this paper, we study the question: \textit{``how can multiple agents be coordinated to asynchronously collaborate over a shared artifact in an effective way?''}

While much research has focused on coordinating multiple agents, ranging from role-based pipelines that mirror human software engineering teams \citep{hong2023metagpt,qian2024chatdev}, to hierarchical managers that decompose and delegate subtasks \citep{benkovich2026agyn}, to include verification mechanisms in multi-agent systems \citep{venkataramani2026mas}, and to automated searches over communication topologies \citep{zhang2025multi}---most of these approaches primarily address how tasks are decomposed and allocated across agents.
However, the core challenges of \emph{asynchronous} multi-agent collaboration over shared artifacts remain unsolved.
Agents operating in this setting face a myriad of challenges such as locally reasonably but globally consistent edits \citep{khatua2026cooperbench}, lack of shared state \citep{cemri2025multi}, and late discovery of any conflicts \citep{cognition2025multiagent}.

Human software engineering teams face these coordination failures routinely, and they have developed a mature infrastructure to mitigate them.
Developers work in isolated copies of the repository (e.g., via \texttt{git worktrees}), so parallel edits do not overwrite one another.
When changes are ready, version-control integration protocols (e.g., merge-based workflows) consolidate contributions and conflicts explicitly, rather than allowing silent interference. Test suites verify each change automatically, so correctness does not rely solely on any single developer's judgment. 


With this in mind, we build \ourmethod (\autoref{fig:teaser}), a multi-agent system grounded in SWE primitives, in which a manager agent dynamically decomposes and delegates tasks to multiple engineer agents who execute concurrently in isolated workspaces.
In particular, each engineer operates in its own \texttt{git worktree}, a fully isolated workspace with a versioned copy of the repository and when an engineer finishes, its changes are integrated back through \texttt{git merge}.
As in human software teams, each engineer is responsible not only for implementation, but also for executable self-verification and conflict resolution at commit time.
Communication between the manager and engineers uses structured \texttt{JSON} instructions and git commits rather than free-form dialog, avoiding the inter-agent misalignment that has been identified as the primary failure mode in multi-agent systems~\citep{cemri2025multi}.
We provide further details on the design of \ourmethod in \autoref{sec:methods}.

We evaluate \ourmethod on two long-horizon, complex software engineering tasks that provide a natural testbed for \emph{shared-artifact} collaboration. Specifically, we use Commit0~\citep{zhao2024commit0}, which requires agents to implement Python libraries from scratch (e.g., \texttt{tinydb}, \texttt{minitorch}, \texttt{jinja}), and on PaperBench~\citep{starace2025paperbench}, where agents reproduce a conference paper. Together, these benchmarks allow us to evaluate \ourmethod with the lens of branch-and-merge coordination in long-horizon multi-agent software engineering. Based on these experiments, we show that \ourmethod consistently improves the performance of Commit0 and PaperBench across multiple models.

\section{Branch-and-Merge Multi-Agent Coordination with SWE Primitives}
\label{sec:methods}

\begin{table}[t]
\centering
\footnotesize
\resizebox{\columnwidth}{!}{
\begin{tabularx}{\columnwidth}{l l X}
\toprule
\textbf{SWE Primitive} & \textbf{Coordination Mechanism} & \textbf{Role in \textsc{Caid}} \\
\midrule
Dependency graph & Scheduling constraints & Dependency order determines safe task delegation \\
\texttt{git worktree} & Workspace isolation & Each agent works in an independent worktree\\
\texttt{git commit} / \texttt{git pull} request & Structured signaling & Agents report completion by making the commits \\
\texttt{git merge} & Output integration & Completed changes are merged into the main\\
Merge conflict resolution & Conflict handling & Engineer resolves integration conflicts by themselves \\
Code review & Verification & Engineer does the self-verification\\
\texttt{asyncio} parallel execution & Concurrent execution & Multiple agents run concurrently \\
Event loop + \texttt{await} & Coordination cycle & Await completion $\rightarrow$ integrate $\rightarrow$ reassign tasks\\
\texttt{git reset $--$hard HEAD} & State synchronization & Worktrees sync to latest integrated state \\
\bottomrule
\end{tabularx}
}
\caption{\textbf{Mapping between concrete SWE primitives and multi-agent coordination mechanisms in \ourmethod.} Each primitive serves as an operational building block for isolation, delegation, asynchronous execution, and integration.}
\label{tab:swe-primitive-mapping}
\end{table}
\ourmethod's coordination architecture is based on SWE primitives, which support operations such as task decomposition, isolated development, integration, and verification. In \autoref{tab:swe-primitive-mapping}, we associate concrete SWE primitives (e.g., \texttt{git worktree}, \texttt{git merge}, dependency graphs, and test suites) and their corresponding coordination roles in \ourmethod.
\ourmethod consists of task specification and dependency modeling (\autoref{sec:task-spec-dep-graph}), dependency-aware task delegation (\autoref{sec:dep-awa-delegation}), workspace isolation and integration (\autoref{sec:iso-int}), structured communication with asynchronous execution (\autoref{sec:com-async}), and self-verification with termination control (\autoref{sec:self-verfication}).

\subsection{Task Specification and Dependency Graph}
\label{sec:task-spec-dep-graph}
To perform multi-agent delegation, we need to split the overall task into ordered sub-tasks. In our preliminary experience, allowing agents to split tasks arbitrarily causes them to miss important parts as they proceed. 
Therefore, to proceed with task delegation in a structured way, we instead have the manager create a dependency graph of the repository to organize the work to be done. The repository structure is represented as a directed graph $G = (V, E)$, where each node $v \in V$ corresponds to a unit of work and each directed edge $(v_i, v_j) \in E$ indicates that $v_j$ depends on $v_i$.
Let $\mathcal{C}_t \subseteq V$ denote the set of units that have been completed and successfully integrated into the main branch at round $t$. A unit $v_j$ is eligible for delegation only if all its dependencies have been satisfied:
$\texttt{Ready}_t(v_j) \iff \forall (v_i, v_j) \in E,\; v_i \in \mathcal{C}_t$.
At each round, the manager selects executable units from the ready set $\{ v \in V \mid \texttt{Ready}_t(v) \}$ and converts them into task assignments.
Depending on the task, the unit of work and dependency analysis method is defined differently. In \autoref{sec:commit0}, we describe these definitions for Commit0 and PaperBench respectively. Although granularity differs across benchmarks, in both settings the manager constructs a dependency structure before delegating, and engineers are assigned tasks only after it is established.

\subsection{Dependency-Aware Task Delegation}
\label{sec:dep-awa-delegation}

We prompt (see Appendix~\ref{app:commit0-prompt} and \ref{app:paperbench-prompt}) the manager to convert the dependency structure from Section~\ref{sec:task-spec-dep-graph} into small executable task units assigned to each engineer. The manager splits implementation work into at most $N$ major task groups, where $N$ is the maximum number of parallel engineers, activating up to $N$ engineers whose dependencies are satisfied (not all $N$ are necessarily activated). Files with strong or circular dependencies are grouped together and assigned to the same engineer to reduce cross-agent coordination.

At each delegation step, the manager selects next tasks with top priority from the major task group, prioritized based on tasks that enable earlier test execution, expose more evaluation signals, or lie closer to the upstream end of the dependency chain. We suggest to the manager that engineers typically start with simpler functions before moving on to more complex ones. The manager dynamically updates dependency state after an engineer merges code and decides whether to assign the next task or keep the engineer idle. We define one round as a complete cycle of delegation, implementation, and dependency update. The process continues until no executable task groups remain or execution limits are reached.

\subsection{Workspace Isolation and Integration}
\label{sec:iso-int}
We use \texttt{git worktree} to ensure that each engineer modifies files only within its workspace, which is always derived from the main branch. Before delegation, we ask the manager to set up the repository in an executable state, including preparing the runtime environment, organizing entry points, or adding minimal function stubs when required by the task. These preparatory changes are committed to the main branch so that all subsequent engineer branches are created from a consistent base state. Certain shared files, such as package initialization files (e.g., \texttt{\_\_init\_\_.py}), are marked as restricted, and engineers are explicitly instructed not to commit changes to them. Worktrees are deleted after all assigned tasks are completed or when the engineer reaches the predefined iteration limit.
Integration is performed through standard \texttt{git commit} and \texttt{git merge} operations. After completing implementation and self-verification, an engineer submits a commit from its branch. The manager attempts to merge this branch into the main branch. If a merge conflict occurs, the engineer who produced the conflicting commit is responsible for resolving it. To solve the conflict, we ask the engineer to pull the latest main branch into its worktree, resolve conflicts locally, and resubmit the updated commit. As a result, the main branch remains the single source of integrated state throughout execution. We observe that this branch-based isolation, combined with explicit merge responsibilities, prevents parallel development from corrupting the shared codebase.

\subsection{Communication and Asynchronous Execution}
\label{sec:com-async}
We use a structured JSON protocol as the communication interface between the manager and the engineer agents. When delegated the task, the manager outputs a JSON specification that defines task assignments, file paths, target functions, and dependency information to ensure that the task boundaries, responsibilities, and outputs are explicitly defined and can be programmatically validated. We provide the details in Appendix \ref{app:commit0-prompt}.

The execution is organized around an asynchronous manager-controlled event loop. Once tasks are delegated, each engineer operates as an independent coroutine. Engineers invoke language model calls, modify code in their worktrees, and execute verification commands such as running tests. These operations are executed concurrently up to a predefined maximum number of active engineers. The manager listens for completion signals and dynamically updates the dependency state when commits are submitted. Engineers who finish early can be assigned new executable task units, while engineers whose dependencies are not yet satisfied remain idle. To manage context growth, the manager maintains a compressed execution history. We use \texttt{LLMSummarizingCondenser} to periodically summarize prior interaction rounds while preserving key structured artifacts such as the dependency graph, completed tasks, and unresolved errors. This separation prevents unnecessary context expansion while preserving execution traceability.

\subsection{Self-Verification and Termination}
\label{sec:self-verfication}
To ensure implementation quality, we require each engineer to perform verification before submitting a commit. If executable tests are available, the engineer runs the subset of tests that directly import or reference the modified files. If there is no explicit mapping, the engineer runs the repository’s default test command or a minimally runnable entry point. Any failed test or runtime exception must be resolved before submission, and engineers iteratively refine the implementation using concrete error logs and tracebacks. After a verified commit is submitted, the manager integrates it into the main branch and updates the dependency state. The manager does not perform a detailed code review at every step, but monitors the overall progress and remaining implementation units. We terminate execution when all units in the dependency structure have been completed and integrated, or when predefined limits, such as maximum rounds or iteration budgets, are reached.

\section{Main Results}
\label{sec:main-results}
\subsection{Evaluation Benchmarks}
We evaluate \ourmethod on two long-horizon software engineering benchmarks.

\label{sec:commit0}
\paragraph{Commit0} \citep{zhao2024commit0} tests whether agents can implement a Python library from scratch given a repository skeleton and a suite of unit tests. The task is considered successful only if all tests pass, making it a repository-level integration problem rather than a collection of independent code completions.
We use Commit0-Lite as our primary evaluation set.
In Commit0, the manager receives an instruction and a repository path containing executable tests (Appendix \ref{app:commit0-prompt}). It first checks import statements for file-level dependencies, collects test cases, and examines which files those tests exercise. The manager is instructed to first consider file-level delegation, but if a single file contains many unimplemented functions it can further divide work at the function level. After assigning initial tasks to multiple engineers, the manager continues exploring the repository until one engineer completes its tasks, submits a commit for merge, and is ready for the next task.

\label{sec:paperbench}
\paragraph{PaperBench} \citep{starace2025paperbench} evaluates an agent's ability to reproduce the main contributions of a published conference paper, typically involving multi-step implementation, experimental setup, and result verification. The benchmark emphasizes long-horizon reasoning and structured execution over complex codebases. Due to computational cost constraints, we adopt the \texttt{Code-Dev} evaluation protocol instead of running the full evaluation pipeline. Following the benchmark’s evaluation paradigm, we use \texttt{gpt-5-mini} \citep{openai2025gpt5mini} as the judge model to assess functional correctness and completion quality. As an open-ended task, explicit test-to-file mappings are not always available. The manager reads the paper, considers the main contribution as the central implementation objective, and infers the required implementation order from it. We provide the prompt in Appendix~\ref{app:paperbench-prompt}.

\subsection{Experimental Setup}
We build \ourmethod using the open-source OpenHands agent SDK~\citep{wang2024openhands,wang2025openhands} (v1.11.0), instantiating a centralized manager for dependency-aware task delegation and multiple software-engineer agents in isolated workspaces. We evaluate with three language models: two open-source (\texttt{GLM 4.7}~\citep{zeng2025glm} and \texttt{MiniMax 2.5}~\citep{minimax2024minimax25}) and one closed-source (\texttt{Claude-4.5-Sonnet}~\citep{anthropic2024claude45}).
Following the Commit0 leaderboard\footnote{\url{https://commit-0.github.io/}} configuration, we use a single-agent setup with $\texttt{max\_iterations}=100$ on both Commit0 and PaperBench. For multi-agent runs, we set $\texttt{max\_iterations}=50$ for the manager and $\texttt{max\_iterations}=80$ for each engineer agent, with $2$ implementation rounds. In the main results, we use one manager with $2$ engineers on PaperBench and $4$ on Commit0. Detailed analysis of configuration choices is in Section~\ref{sec:analysis}.%
\footnote{All configurations are fixed prior to experimentation to balance correctness and runtime efficiency.}


\subsection{Baselines}
Our primary baseline is a single-agent system built on the same OpenHands agent, isolating the effect of branch-and-merge coordination while holding the underlying framework fixed. This controlled comparison measures the incremental contribution of dependency-aware delegation, isolated workspaces, and merge-and-branch integration without introducing variation from framework-level differences such as prompting structure, tool interfaces, memory mechanisms, or execution policies. In Section~\ref{sec:analysis}, we further vary coordination and isolation mechanisms to compare different multi-agent architecture design choices.


\subsection{Branch-and-Merge Based Coordination Improves Multi-Agent Performance}
\begin{table}[h]
\small
\centering
\resizebox{\columnwidth}{!}{
\begin{tabular}{l c ccc ccc ccc}
\toprule
\midrule
\rowcolor{gray!15}
\multicolumn{11}{c}{\textbf{PaperBench}} \\
\midrule
& & \multicolumn{3}{c}{\textbf{Single-Agent}}
    & \multicolumn{3}{c}{\textbf{\textsc{\ourmethod} (2 Engineers)}}
    & \multicolumn{3}{c}{\textbf{\textsc{Single-Agent + \ourmethod}}} \\
\cmidrule(lr){3-5} \cmidrule(lr){6-8} \cmidrule(lr){9-11}
\textbf{Model} & \textbf{SDK}
  & Score & Runtime & Cost
  & Score & Runtime & Cost
  & Score & Runtime & Cost \\
\midrule
Claude Sonnet 4.5 & v1.11.0 & 57.2 & 1803.5 & 3.3 & 63.3 & 2080.4 & 6.5 & \textbf{66.8} & 3883.9 & 9.7 \\
MiniMax 2.5         & v1.11.0 & 10.5   & 2525.3    & 1.1  & 36.1 & 3042.4 & 2.6 & \textbf{36.7} & 5567.7 & 3.7 \\
GLM 4.7           & v1.11.0 & 38.0 & 1177.6 & 2.8   & 45.4 & 1449.4 & 4.7 & \textbf{48.5} & 2627.0 & 7.5 \\
\midrule
\rowcolor{gray!15}
\multicolumn{11}{c}{\textbf{Commit0-Lite}} \\
\midrule
& & \multicolumn{3}{c}{\textbf{Single-Agent}}
    & \multicolumn{3}{c}{\textbf{\textsc{\ourmethod} (4 Engineers)}}
    & \multicolumn{3}{c}{\textbf{\textsc{Single-Agent + \ourmethod}}} \\
\cmidrule(lr){3-5} \cmidrule(lr){6-8} \cmidrule(lr){9-11}
\textbf{Model} & \textbf{SDK}
  & Score & Runtime & Cost
  & Score & Runtime & Cost
  & Score & Runtime & Cost \\
\midrule
Claude Sonnet 4.5 & v1.11.0 & 53.1 & 692.6 & 1.9 & 59.1 & 1583.2 & 8.1 & \textbf{59.5} & 2275.8 & 10.0 \\
MiniMax 2.5       & v1.11.0 & 42.3 & 752.1 & 1.6 & \textbf{57.0} & 1908.7 & 4.5 & 57.0 & 2660.7 & 6.2 \\
GLM 4.7           & v1.11.0 & 42.9 & 871.0 & 2.5 & \textbf{46.5} & 1387.8 & 7.3 & 46.5 & 2258.8 & 9.8 \\
\midrule
\bottomrule
\end{tabular}
}
\caption{\textbf{Main results on Commit0 and PaperBench.} We compare single-agent baselines with \ourmethod (2 engineers on PaperBench and 4 engineers on Commit0) under the same underlying model and fixed per-configuration iteration budgets.}
\label{tab:main-results}
\end{table}

We compare \ourmethod with the single-agent baseline in Table \ref{tab:main-results} and observe a consistent advantage for the branch-and-merge-based multi-agent system across both benchmarks and three LLMs. On PaperBench, we observe that multi-agent coordination yields large gains for weaker single-agent runs: MiniMax 2.5 reaches 36.1\% under multi-agent execution, while its single-agent score is only 10.5\%. The improvement is not limited to weaker models. With Claude 4.5, multi-agent execution achieves 63.3\% compared to 57.2\% for single-agent. In Commit0-Lite, we find the same pattern. Claude 4.5 improves from 53.1\% to 59.1\%, and MiniMax 2.5 reaches 57.0\% under multi-agent execution. These results indicate that the performance gap is not explained by changing the underlying model, but by changing the execution method. In \ourmethod, engineers work in separate branches and changes enter the main branch only through explicit merge and test validation. This makes parallel work usable by separating implementation from integration: engineers can iterate locally without overwriting each other’s intermediate states, while integration failures are surfaced at merge time with concrete test signals tied to specific updates.  Our results in Table \ref{tab:main-results} are consistent with the benefit of making integration explicit and test-gated under long-horizon execution. We provide one-sided t-tests in Appendix \ref{app:stats-test}.

Table \ref{tab:main-results} further reveals an important strategic implication. In long-horizon shared-artifact tasks, multi-agent coordination should not be treated as a fallback after single-agent failure. The Single-Agent + Multi-Agent setting approximates a practical strategy in which a single agent is first attempted, followed by coordinated execution if necessary. However, this sequential strategy incurs nearly additive runtime and cost, while the final performance remains close to the direct multi-agent result. For example, on PaperBench with Claude Sonnet 4.5, the combined strategy reaches 66.8\%, only slightly above the multi-agent score of 63.3\%, yet runtime increases from 2080.4s to 3883.9s and cost rises from 6.5 to 9.7. On Commit0-Lite with MiniMax 2.5, the multi-agent score is 57.0\%, and the combined strategy remains 57.0\%, while both runtime and cost increase substantially. These results give us a clear strategy insight for long-horizon shared-artifact tasks. 
Treating multi-agent coordination as a fallback after a single-agent attempt is inefficient. A more cost-effective strategy is to adopt coordinated multi-agent execution from the outset rather than switching only after failure.

\subsection{Single Agents Fail to Utilize More Iterations}


\begin{figure}[h]
    \centering
    \small
    \includegraphics[width=\columnwidth]{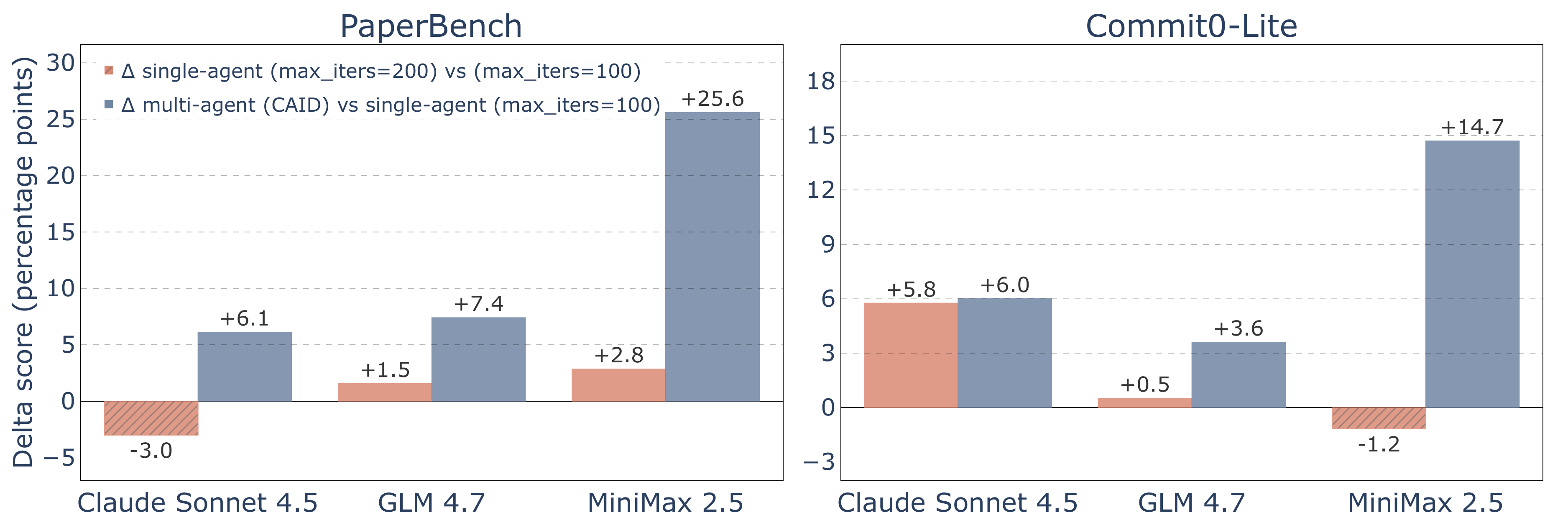}
    \caption{\textbf{\ourmethod effectively utilizes iteration budgets.} We compare the final score and the iteration utilization between single-agent runs with different iteration limits and \ourmethod.
}
    \label{fig:single-agent-iter}
\end{figure}

\textbf{\textit{Can a single agent overcome long-horizon shared-artifact challenges simply by running longer?}} 
To study this, we run a single agent with $max\_iterations=100$ and $max\_iterations=200$. We control computation through a max iteration budget rather than enforcing a fixed runtime, which better reflects practical agent deployment where iteration-based control is commonly used.
As shown in Figure~\ref{fig:single-agent-iter}, doubling the iteration limit yields only marginal improvements and, in some cases, even degraded results. In PaperBench, $\Delta$ from 100 to 200 iterations remains small for GLM 4.7 and MiniMax 2.5, and becomes negative for Claude Sonnet 4.5. In Commit0-Lite, the improvement is similarly limited, and MiniMax 2.5 shows a negative delta. This trend is consistent with the findings in PaperBench, where forcing the agent to run until a time limit does not reliably improve the judge score~\citep{starace2025paperbench}. In Figure~\ref{fig:single-agent-iter}, we also show the score gain of \ourmethod relative to the 100-iteration single-agent baseline. Across both benchmarks, these gains are substantially larger than those from increasing the iteration budget. For example, on PaperBench the multi-agent improvement for MiniMax 2.5 exceeds 25 percentage points, while doubling iterations yields only a small change. A similar gap appears in Commit0-Lite. These results show that extending the iteration budget alone does not resolve the fundamental bottleneck of a single agent on long-horizon tasks, whereas multi-agent coordination produces significantly larger gains.

\section{Analysis}
\label{sec:analysis}
\subsection{\texttt{Git worktree} Isolation}
\begin{wraptable}{r}{0.48\columnwidth}
\vspace{-0.5em}
\centering
\small
\setlength{\tabcolsep}{3pt}
\renewcommand{\arraystretch}{1.1}

\begin{tabular}{cc cc cc}
  \toprule
  \rowcolor{gray!15}
  \multicolumn{6}{c}{\textbf{PaperBench}} \\
  \midrule
  \multicolumn{2}{c}{single agent} &
  \multicolumn{2}{c}{\makecell{\ourmethod\\(worktree isolation)}} &
  \multicolumn{2}{c}{\makecell{multi-agent\\(soft isolation)}} \\
  \cmidrule(lr){1-2}\cmidrule(lr){3-4}\cmidrule(lr){5-6}
  score & iterations & score & iterations & score & iterations \\
  \midrule
  57.2 & 66.8 & 63.3 & 168.3 & {\color{red}\textbf{55.5}} & 190.0 \\
  \midrule
  \rowcolor{gray!15}
  \multicolumn{6}{c}{\textbf{Commit0-Lite}} \\
  \midrule
  \multicolumn{2}{c}{single agent} &
  \multicolumn{2}{c}{\makecell{\ourmethod \\(worktree isolation)}} &
  \multicolumn{2}{c}{\makecell{multi-agent \\ (soft isolation)}} \\
  \cmidrule(lr){1-2}\cmidrule(lr){3-4}\cmidrule(lr){5-6}
  score & iterations & score & iterations & score & iterations \\
  \midrule
  {\color{red}\textbf{53.1}} & 84.5 & 59.1 & 313.3 & 56.1 & 335.9 \\
  \bottomrule
\end{tabular}

\caption{We compare soft context isolation and worktree isolation on PaperBench and Commit0-Lite.}
\label{tab:isolation}
\end{wraptable}

In Table \ref{tab:isolation}, we study whether our proposed method of ``worktree isolation'' is necessary, comparing it with ``soft isolation'', where all engineers share one workspace, and the central manager attempts to prevent conflicts through instruction-level constraints, such as assigning non-overlapping files and explicitly warning against interference.
On Commit0-Lite, soft isolation improves over single-agent from 53.1\% to 56.1\%, showing that central manager-driven delegation alone already helps when repository structure and file dependencies are explicit. Worktree isolation further increases performance to 59.1\%, indicating that instruction-level separation is not sufficient to fully eliminate interference over longer trajectories. In contrast, on PaperBench soft isolation drops to 55.5\%, below the single-agent score of 57.2\%, while worktree isolation reaches 63.3\%. Unlike Commit0, PaperBench does not provide explicit file structure or dependency graphs, and the manager must first infer the global implementation plan from the paper itself. In this case, sharing a workspace causes miscoordination, whereas worktree isolation stabilizes parallel execution.

\subsection{Choosing the Degree of Parallel Execution}
\begin{figure}[h]
    \centering
    \small
    \includegraphics[width=0.9\columnwidth]{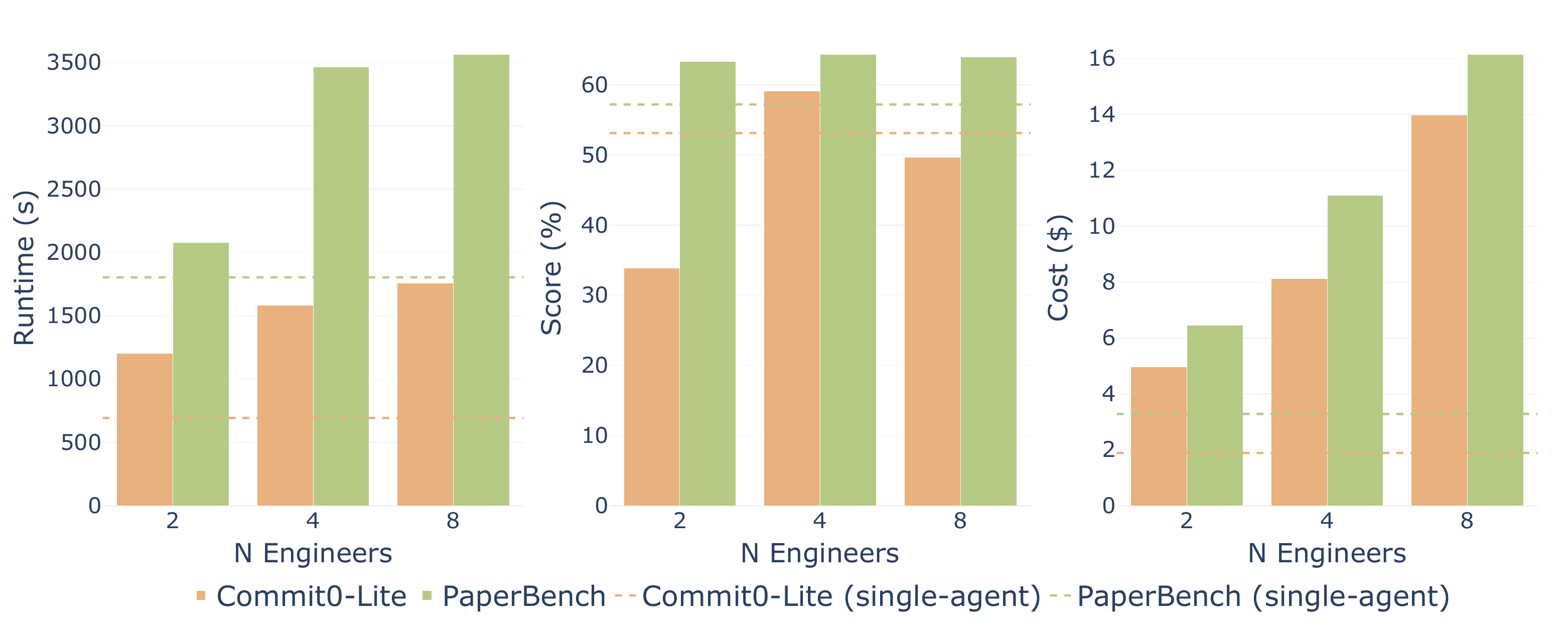}
    \caption{\textbf{Effect of the number of engineer agents on runtime, pass rate, and cost for Commit0-Lite and PaperBench.} We provide the single-agent baselines here for comparison.}
    \label{fig:num-subagents}
\end{figure}
We analyze how the number of asynchronous engineer agents affects the performance in Figure \ref{fig:num-subagents}. We find that increasing the number of engineers does not monotonically improve the performance, which aligns with the results in \citep{yang2026understanding}. The optimal degree of parallelism depends on two factors: the intrinsic parallel structure of the task and the delegation capacity of the central manager. First, tasks differ in how many components can be implemented independently. In Commit0-Lite, performance improves when increasing engineers from 2 to 4, but decreases when expanding to 8 engineers. Although more agents increase theoretical parallelism, overly fine-grained task delegation introduces integration overhead and conflict resolution cost, especially when multiple engineers modify closely related modules. However, too few engineers can exploit the independent files available in clear-structured repositories, limiting progress within a fixed iteration budget. Second, scalability is constrained by the manager's coordination ability. The central manager must track dependency states, monitor the progress of engineers, and dynamically assign tasks. When the number of engineers increases, delegation errors or delayed synchronization can propagate and destabilize the overall trajectory. This effect is visible in Commit0-Lite at 8 engineers, where performance declines despite higher computation cost. On PaperBench, where task decomposition is less structurally explicit, increasing engineers beyond 2 yields minimal gain in score while runtime and cost increase steadily. These results show that the number of subagents should be matched to both the inherent modularity of the task and the effective delegation capacity of the manager. Excess parallelism without reliable coordination degrades stability, rather than improving performance. We provide examples of failure in the Appendix \ref{app:failure-parallel}.

\subsection{Delegation Shapes Execution Trajectory}
\label{sec:delegation}
\begin{figure*}[h!]
    \centering
    \includegraphics[width=0.95\textwidth]{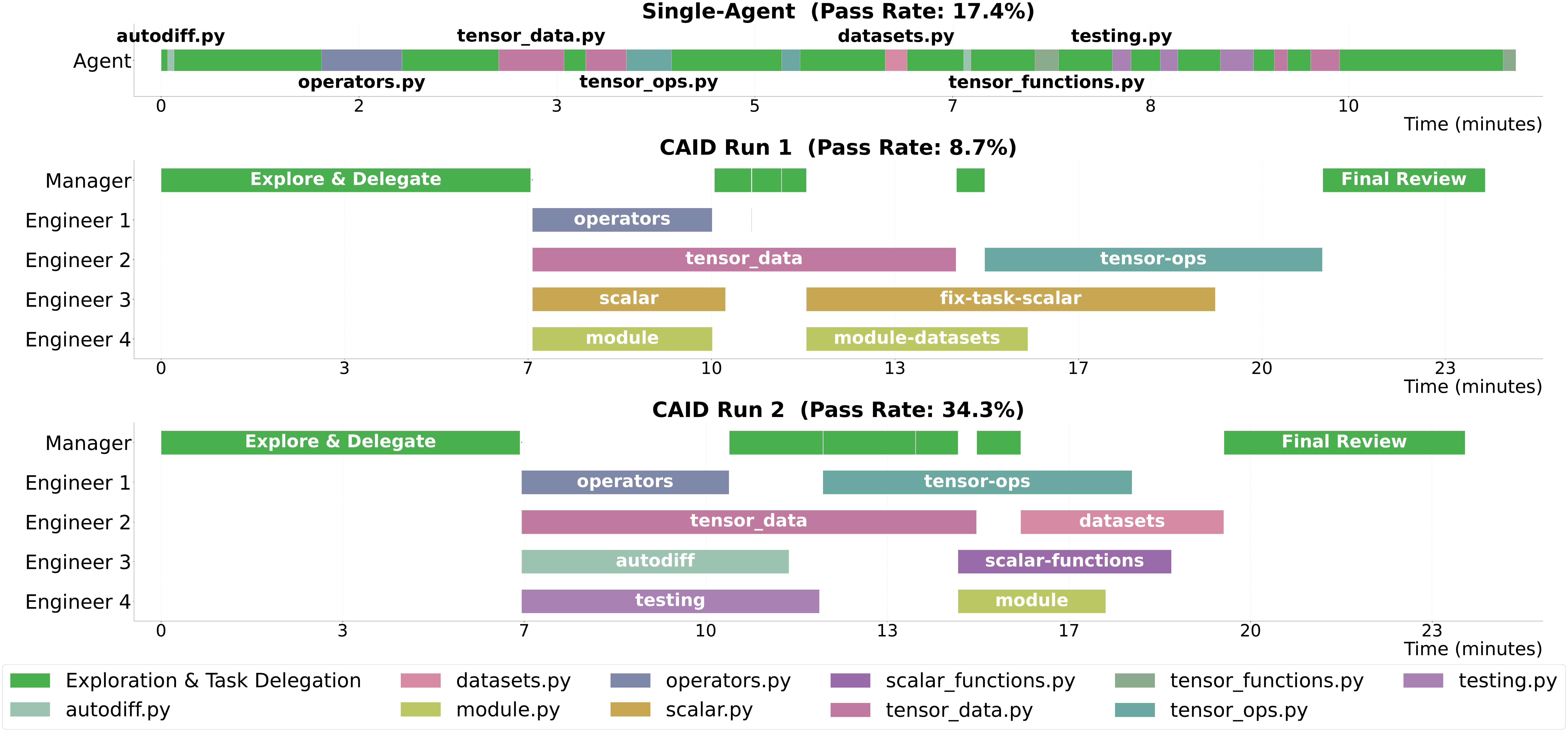}
    \caption{\textbf{Execution timelines on the \texttt{minitorch} repository for a single-agent run and two \ourmethod runs.} The bars in the Gantt plot indicate file-level implementation intervals and manager phases. The runs differ in which modules are assigned and actively developed, resulting in distinct execution trajectories and pass rates.}
    \label{fig:parallel-execution}
\end{figure*}
In Figure \ref{fig:parallel-execution}, we show two \ourmethod runs and one single-agent run in the \texttt{Commit0-Lite minitorch} repository to study how task delegation affects execution outcomes. We find that the performance difference between \ourmethod Run 1 (8.7\% pass rate) and \ourmethod Run 2 (34.3\%) is not simply due to the number of modules implemented, but to which modules are assigned and actively pursued. In Run 2, the manager assigns an engineer to \texttt{autodiff.py}, a file that is critical for passing tests, and sustained effort on this file is followed by broader progress across dependent components. In contrast, Run 1 assigns engineers to several other files, but never assigns work to \texttt{autodiff.py}. Although multiple engineers are active, the absence of this key dependency limits the overall pass rate. We observe that the single-agent run touches \texttt{autodiff.py} during exploration and implements part of the logic, but the file remains incomplete and the final pass rate reaches only 17.4\%. This example shows that the manager's delegation ability, particularly the ability to identify and assign high-impact dependencies, is critical for the success of long-horizon SWE tasks.

\subsection{Scaling Asynchronous Parallelism}

\begin{wrapfigure}{l}{0.5\columnwidth}
\vspace{-1.8em}
\centering
\includegraphics[width=\linewidth]{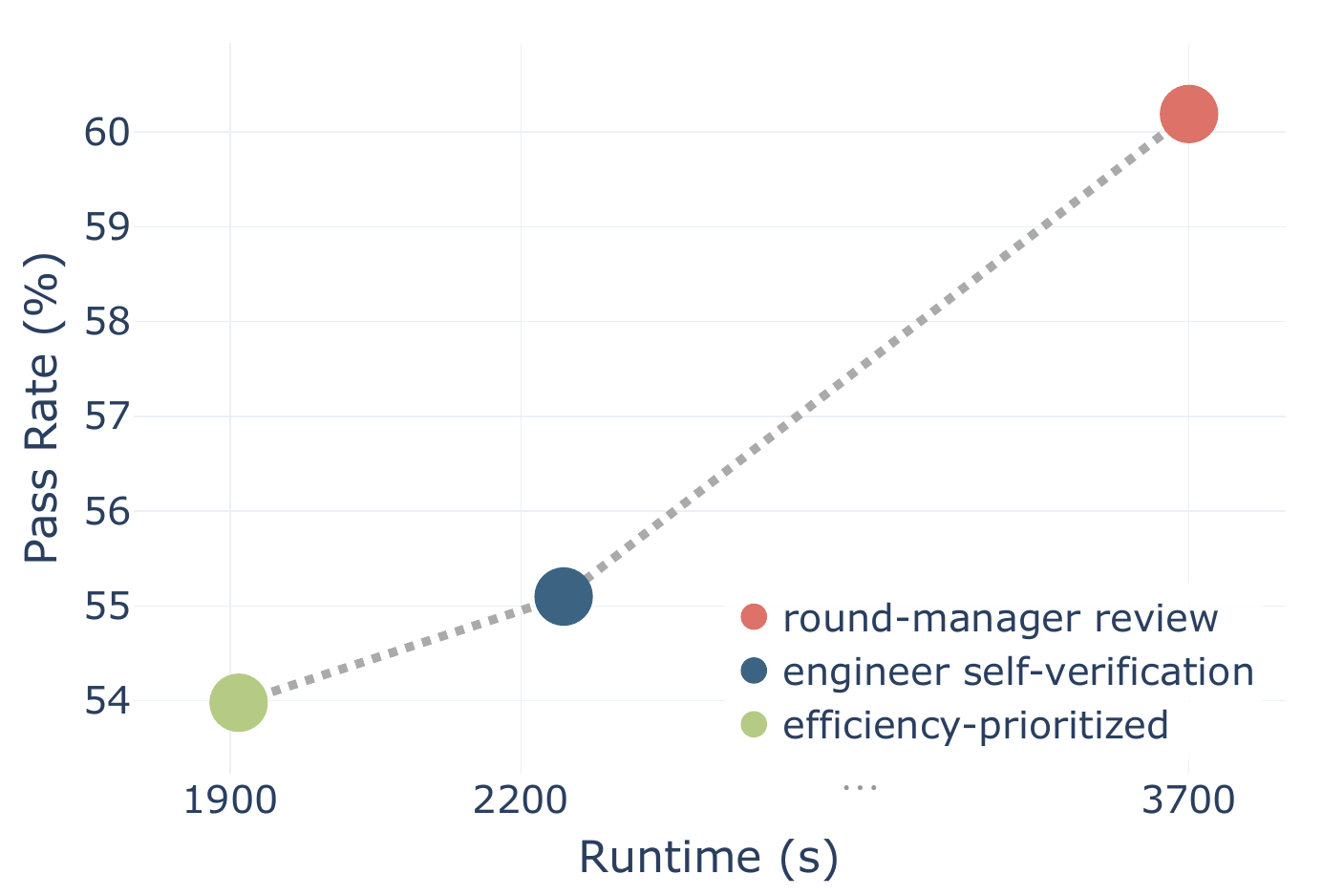}
\caption{\textbf{Runtime (s) vs. pass rate (\%) of a subset of the Commit0 under three coordination prompts} (1) Round-Manager Review: the manager reviews each round before integration; (2) Engineer Self-Verification: engineers verify locally without repeated managerial review; and (3) Efficiency-Prioritized: all agents are instructed to prioritize runtime efficiency.}
\label{fig:task-delegation}
\vspace{-1em}
\end{wrapfigure}

During our exploration of the multi-agent design, we experimented with different prompt engineering strategies that emphasize distinct objectives, such as prioritizing correctness or efficiency. Figure \ref{fig:task-delegation} shows the results on a subset of eight repositories (i.e., \emph{babel, chardet, cookiecutter, imapclient, jinja, minitorch, simpy, tinydb}) of Commit0-Lite. In \textit{Round-Manager Review}, the manager explicitly reviews code quality at every implementation round before integration for each engineer, placing stronger emphasis on correctness. In \textit{Engineer Self-Verification}, engineers conduct self-review without repeated managerial inspection, which is closest to the main results we report in Section \ref{sec:main-results}. In \textit{Efficiency-Prioritized}, both manager and engineer agents are explicitly instructed to prioritize runtime efficiency and are reminded that execution time will be evaluated, thereby assigning higher weight to the runtime of implementation in the user instruction. We observe a clear pattern: \textit{Round-Manager Review} achieves the highest pass rate (60.2\%) but also incurs the longest runtime (3689.1s), \textit{Self-Verification} yields intermediate performance (55.1\%) with moderate runtime (2243.9s), and \textit{Efficiency-Prioritized} runs fastest (1908.6s) but achieves the lowest pass rate (54.0\%). This development-stage result suggests a trade-off between verification intensity and execution efficiency: emphasizing efficiency can shorten runtime but may reduce integration robustness, while stricter review improves stability at additional computational cost.








\section{Related Work}
\subsection{Multi-Agent Architectures}
Recent studies have explored diverse architectural choices for LLM-based multi-agent systems, spanning from static, predefined role-playing topologies to dynamic, task-adaptive orchestrations. Early frameworks such as CAMEL \citep{li2023camel} and Generative Agents \citep{park2023generative} established the foundation for communicative interaction, which was later structured into a natural language communication pipeline from ChatDev \citep{qian2024chatdev}. To enhance flexibility, EvoMAC \cite{hu2024self} 
explores self-evolving collaboration and AutoAgents \cite{chen2023autoagents} focuses on automated agent generation. Advanced orchestrators like AgentOrchestra \cite{zhang2025agentorchestra} introduce standardized protocols (e.g., TEA), while MASS \citep{zhou2025multi} and DyLAN \citep{liu2023dynamic} optimize inter-agent topologies for adaptive task decomposition. Despite achieving higher autonomy in personnel allocation, these architectures still struggle with high-density communication and cognitive overload in long-horizon tasks. To address this, MegaAgent \cite{wang2025megaagent} and subsequent scaling laws \citep{qian2024scaling} examine the decay of efficiency in large clusters, leading to optimization strategies such as sequential aggregation in Chain-of-Agents \citep{zhang2024chain}, and memory abstractions in MemGPT \citep{packer2023memgpt}. Many open-source agents such as OpenHands \citep{wang2024openhands} further reduce context explosion through history condensation.

Although many multi-agent systems optimize information flow, they largely rely on "standardized operating procedures" to maintain agent coordination \citep{hong2023metagpt, nguyen2025agilecoder} and incorporate agile methodologies for lifecycle management. Deeper coordination is studied through implicit co-player inference \citep{meulemans2024multi}, consensus-based evaluation in agent-as-judge \citep{zhuge2024agent}. However, in shared-artifact environments like software engineering, these linguistically-governed architectures frequently encounter execution conflicts when multiple agents concurrently modify the codebase. \cite{khatua2026cooperbench} suggest that this critical bottleneck for multi-agent execution remains under-explored. This gap reveals that we need an architectural design that physically coordinates multiple agents in an execution-aware paradigm.

\subsection{Multi-Agent Coordination Challenges}
Despite advances in multi-agent architectures, coordination stability remains constrained by communication workflows, which is directly reflected in task delegation under the uncertainty of complex tasks and explicit conflicts within shared workspaces. In dialogue-driven systems \citep{wu2024autogen}, delegation typically emerges implicitly through conversational interaction rather than explicit authority modeling, which can lead to redundant effort or delayed escalation. While recent studies propose more structured approaches—including orchestrator-executor handoffs and hierarchical organizations \citep{song2025coact, xu2025boad} to regulate task delegation, scaling analyses \citep{qian2024scaling, li2024more} demonstrate that increasing the agent population without disciplined delegation amplifies communication overhead and may degrade overall performance. Another critical challenge caused by unstructured communication is physical interference: planning-oriented analyses \citep{li2024agent} report severe task overlap and inconsistent action sequences, while empirical scaling results \citep{qian2024scaling, li2024more} quantify a "coordination tax" in which synchronization costs grow superlinearly with agent count. These findings indicate that linguistic alignment can harmonize intent but cannot inherently serialize concurrent state transitions or guarantee integration consistency. To address these challenges, we use a central manager to explicitly delegate tasks and physically isolate the workspaces of concurrent agents to prevent integration conflicts.

\subsection{Software Engineering for Multi-Agent Coordination}
Before the emergence of LLM-based agents, software engineering had already developed mechanisms for coordinating parallel work over shared artifacts, including branching and merging, dependency management, continuous integration, and code review. These mechanisms treat coordination as explicit control over versioned artifacts and their integration. Recent multi-agent work has begun to implicitly adopt parts of the SWE paradigm. Process-driven frameworks such as MetaGPT \citep{hong2023metagpt} and AgileCoder \citep{nguyen2025agilecoder} mirror role decomposition and lifecycle management. Sandbox-based systems, including the SWE-agent \cite{yang2024swe}, incorporate build–test feedback loops analogous to continuous integration. However, recent empirical studies \cite{khatua2026cooperbench} still report that concurrent modification and merge conflicts remain a primary failure mode when these engineering primitives are not explicitly modeled. These observations suggest that, in shared repositories, the central issue is not only how agents are organized into roles or workflows, but also how concurrent work is isolated, integrated, and verified.  In this paper, we focus on branch-and-merge coordination and the SWE primitives that support it in multi-agent software engineering.


\subsection{Software Engineering Evaluation Benchmarks}
Software Engineering (SWE) tasks, which evaluate agents on their ability to autonomously carry out diverse real-world development activities across complex codebases, have become the core benchmarks for measuring the practical capabilities of LLM-based coding agents. SWE-bench \cite{jimenez2023swe} provides the initial benchmark for autonomous issue resolution. SWE-bench Verified \cite{chowdhury2024swebenchverified} refines the evaluation methodology to enhance fidelity and robustness, whereas SWE-bench Pro \cite{deng2025swe} expands the task design to include professionally curated, multi-step engineering problems that better approximate complex real-world development workflows. To move beyond issue-level resolution, several benchmarks isolate specific capabilities of software engineering agents. TerminalBench \cite{merrill2026terminal} and InterCode \cite{yang2023intercode} evaluate the use of terminal-based tools, while DevBench \cite{li2024devbench} extends the assessment to the broader software development lifecycle. For long-horizon and reasoning-intensive scenarios, SciCode \citep{tian2024scicode} and LongCLI \citep{feng2026longcli} introduce multi-step algorithmic or decentralized workflows. At a larger granularity, Commit0 \cite{zhao2024commit0} and PaperBench \cite{starace2025paperbench} introduce long-horizon SWE tasks that move beyond localized reasoning. Long-horizon, complex SWE tasks naturally constitute a rigorous testbed for multi-agent systems, as coordinated multi-file modifications, interdependent subtasks, and explicit merge conflicts systematically expose challenges in synchronization, consistency maintenance, and progress integration across agents. In this paper, we evaluate \ourmethod on Commit0 and PaperBench.

\section{Limitations and Future Directions}
\label{sec:limitations-fur-dir}

\paragraph{Cost and Runtime.}
Although \ourmethod improves success rates on long-horizon shared-artifact tasks, it introduces non-trivial coordination overhead. In our experiments, multi-agent execution consistently incurs higher API cost than single-agent baselines, and wall-clock runtime is not substantially reduced despite parallel execution. This reflects a fundamental trade-off: structured isolation, integration, and verification improve stability, but require additional communication rounds, merge operations, and test executions. In particular, while engineers operate concurrently, integration remains sequential and test-gated, limiting end-to-end acceleration. Prior analyses of multi-agent systems have similarly noted that coordination complexity can offset gains from specialization and parallelism when not carefully optimized~\citep{oreilly2024multiagent}. For the long-horizon shared-artifact tasks we study, however, such coordination may still be necessary, since simply extending single-agent execution does not reliably achieve comparable gains. Therefore, promising next steps include improving scheduling efficiency, reducing redundant verification cycles, and learning when to merge or prune intermediate states. Optimizing the cost–performance frontier of structured multi-agent execution remains an important area for future work. 



\paragraph{Isolated Task Delegation Capabilities of Agents.}
A second limitation lies in the central manager’s task decomposition and delegation capability. In the current implementation, task assignment relies primarily on prompt engineering heuristics rather than learned delegation policies. While our results indicate that architectural isolation and integration are the dominant factors for stability, weak or suboptimal task decomposition can still reduce overall effectiveness. Existing analyses of multi-agent systems identify imprecise task handoffs and underspecified subgoals as major sources of coordination failure~\citep{galileo2026failures}. 
Our findings align with this observation: when delegation is coarse-grained or misaligned with dependency structure, engineers may produce locally correct outputs that are globally inefficient to integrate. Future work may explore reinforcement learning–based delegation policies, dependency-aware planning modules, or adaptive subtask refinement strategies that improve alignment between global objectives and isolated execution. Strengthening delegation capability would allow the architectural benefits of isolation and structured integration to scale more reliably.

\paragraph{Generalization Beyond Software Engineering Tasks.}
Finally, our evaluation focuses on software engineering benchmarks, which provide a natural testbed for structured multi-agent execution due to explicit workspace boundaries, version control infrastructure, and executable test suites. These properties make software development uniquely suitable for studying isolation, integration, and dependency-aware coordination. However, not all long-horizon shared-artifact tasks possess such clearly defined boundaries or objective verification mechanisms. Extending \ourmethod to non-coding domains—such as document synthesis, research planning, or multimodal artifact construction—will require adapting isolation mechanisms and designing alternative forms of integration and validation. Evaluating the framework in such settings is necessary to determine whether the architectural principles demonstrated here generalize beyond SWE-specific workflows.
\section{Conclusion}
\label{sec:conclusion}
In this paper, we introduce \ourmethod, a branch-and-merge based multi-agent system for long-horizon software engineering tasks. We use a manager to break a task into dependency-aware units, assign them to engineers, and keep each engineer working in an isolated branch and worktree. Progress is integrated only through \texttt{git commit} and \texttt{git merge} on the main branch, with tests used as the executable check for whether an update should be kept. Across Commit0 and PaperBench, our \ourmethod consistently improves over single-agent baselines, even when the underlying model is unchanged. Our results also show that simply increasing the single agent iteration budget does not reliably improve outcomes, and a fallback strategy that runs a single agent first and then switches to multi-agent mainly wastes runtime and cost. Overall, we show that branch-and-merge is important for effective multi-agent software engineering and that SWE primitives provide a practical way to support it. For complex long-horizon, dependency-aware software engineering tasks, \ourmethod is the default paradigm for structuring solutions to enable parallel and coordinated development.


\section*{Acknowledgments}
This paper was supported by grants from Fujitsu. We thank Apurva Gandhi, Lintang Sutawika, Emmy Liu, and Howard Chen for their valuable feedback and discussion.

\bibliographystyle{unsrtnat}
\bibliography{ref}

\newpage

\appendix

\section{Prompt Engineering for Multi-Agent Task Delegation}
We provide the user instruction and task delegation prompts for both Commit0 in Section \ref{app:commit0-prompt} and PaperBench in \ref{app:paperbench-prompt}.
\newtcblisting{promptbox}[1]{
  enhanced,
  breakable,
  listing only,
  colback=white,
  colframe=black,
  boxrule=0.8pt,
  arc=1mm,
  left=2mm,
  right=2mm,
  top=2mm,
  bottom=2mm,
  title=\texttt{#1},
  colbacktitle=black,
  coltitle=white,
  fonttitle=\bfseries,
  boxed title style={
    colframe=black,
    colback=black,
    boxrule=0.8pt,
    arc=1mm
  },
  listing engine=listings,
  listing options={
    basicstyle=\ttfamily\scriptsize,
    breaklines=true,
    columns=fullflexible,
    keepspaces=true,
    showstringspaces=false
  }
}

\subsection{Commit0 Prompts}
\label{app:commit0-prompt}

\begin{promptbox}{user instruction}
<uploaded_files>
/workspace/{{ workspace_dir_name }}
</uploaded_files>
I've uploaded a Python code repository in the directory {{ workspace_dir_name }}.

Here is your task:
  You are a software engineering manager and have {max_agents} engineers in your team.
  Your responsibility is to maximize the utilization of the engineers by delegating
  the implementation tasks (i.e., the functions with `pass` statements) to these
  engineers and guide them to efficiently and effectively complete the implementation
  and pass ALL the unit tests. Except for the functions with `pass` statements, the
  repository might also contain some missing functions that are not defined in any
  files. You need to add them with clear docstrings and `pass` statements into the
  files for the engineers to implement. Make sure you submit a local commit of
  your changes. Remember you are NOT allowed to generate any code for the existing
  functions or classes with `pass` statements. You can only add undefined
  functions as needed.

  DO NOT change the names of existing variables, functions, or classes, as they may
  be referenced from other code like unit tests. Do not comment out any existing code.

  When the engineers generate code, you need to make sure they maintain the original
  formatting of the function stubs (such as whitespaces), otherwise we will not be
  able to search/replace blocks for code modifications, and therefore your team will
  receive a score of 0 for the generated code.

Here is the command to run the unit tests:
<test_command>
{test_cmd} {test_dir}
</test_command>

Each engineer is expected to proactively submit a local git commit to you once their
assigned task is complete. The engineers are responsible for verifying their own
implementation quality and running tests before submitting. If no commit is submitted,
you should assume the task may be partially complete. In that case, manually inspect
the engineer's worktree, determine which parts have already been implemented, sync
and merge those completed artifacts into the main directory, and make a commit on
their behalf.

When an engineer has completed their task with a successful commit, you need to
decide the next task to assign, following these steps:
  1. Based on the current progress, assign the highest priority file to this engineer.
  2. Make sure the assigned files do not contain or depend on any missing (undefined) functions; if so, add them with clear docstrings and `pass` statements into the files and submit a local commit of your changes.
  3. Explain the overall progress of the implementation and provide a detailed explanation of the purpose of the implementation to instruct the engineer
     to complete the task.
  4. If no new implementation is needed for now (e.g., the file is already implemented or needs to wait for other engineers to complete the dependencies), you can simply say "Thank you for your work. I will assign a new task to you later."
  5. You can also assign tasks to idle or inactive engineers if you need more capacity to better utilize the engineers.

Engineers are responsible for verifying their own implementation quality before
submitting their commits; you do NOT need to review their code quality or run tests
yourself. Only focus on delegating tasks (i.e., maximizing the utilization of the
engineers to pass more unit tests).

Make sure you DO NOT iteratively overcheck or fix missing functions.
Provide a clear and concise response in the JSON format below.
Please structure your response as JSON:

{
  "assign_task": {
    "reasoning": "Explain your decision",
    "assignments": [
      {
        "engineer_id": "engineer_id",
        "task_id": "task-unique-id or 'fix-<original-task-id>'",
        "file_path": "path/to/file.py",
        "functions_to_implement": ["func1", "func2"],
        "instruction": "Detailed explanation of the purpose of the implementation.",
        "complexity": "simple|medium|complex"
      }
    ]
  }
}

If no tasks should be assigned, use an empty assignments array.
\end{promptbox}

\begin{promptbox}{task delegation}
Your engineers are waiting for your instructions to start their first implementation tasks. Now you need to:

  1. Check for uncommitted changes and commit if needed:
     git status
     git add -A && git commit -m "Add missing stubs from scan phase" || true

  2. Suspend exploration and systematically delegate the implementation work
     by outputting a delegation JSON based on your current understanding
     of the repository structure and its dependencies.
     You have up to {max_agents} engineers available.

Suggestions for effective delegation in the first round:

  - First, divide the overall implementation work into up to {max_agents}
    major tasks, balancing complexity and estimated effort as evenly as possible. Keep highly interdependent files within the same major task. Prefer splitting at the file level. If one file contains a disproportionately large number of functions with pass statements, you may split by function and assign non-overlapping sets to multiple engineers.

  - For each engineer, assign the highest-priority file within their major task. If two files are circularly dependent, assign them to the same engineer. Engineers are generally more comfortable starting from simpler tasks before moving to more complex ones.

  - Engineers are only responsible for implementing functions with pass
    statements. Do not assign them to implement missing functions that are not defined in any file. If you previously added undefined functions with pass statements, include them in the assignment instructions.

  - In each assignment, briefly summarize the relevant repository structure and dependencies so engineers do not need to re-explore the codebase. Clearly specify which file and which functions to implement. Explain the purpose and expected behavior of each function. If assigned functions depend on other stub functions not assigned to the same engineer, provide a short description of those dependencies to avoid confusion.

Note: Do NOT provide any code snippets or pseudo-code.
Output your delegation plan strictly in the following JSON format:

{
  "delegation_plan": {
    "first_round": {
      "num_agents": <integer (1 to {max_agents})>,
      "reasoning": "Explain why these files are assigned first and why this number of engineers is used.",
      "tasks": [
        {
          "engineer_id": "engineer_id",
          "task_id": "task-unique-id",
          "file_path": "path/to/file.py",
          "functions_to_implement": ["func1", "func2"],
          "complexity": "simple|medium|complex",
          "instruction": "Summarize the repository structure and dependencies. Then provide detailed instructions for the implementation, including the expected behavior of the assigned functions and descriptions of any dependent stub functions."
        }
      ]
    },
    "remaining_tasks": [
      {
        "task_id": "task-unique-id",
        "file_path": "path/to/file.py",
        "functions_to_implement": ["func1", "func2"],
        "complexity": "simple|medium|complex",
        "depends_on": ["file_path_1", "file_path_2"]
      }
    ]
  }
}
\end{promptbox}

\subsection{PaperBench Prompts}
\label{app:paperbench-prompt}

\begin{promptbox}{user instruction}
You are tasked with reproducing a research paper.

You have been provided with a research paper located at `/workspace/paper`. The directory contains the paper in both PDF and markdown format.

Here is your task:
You are a software engineering manager for a research team and have {max_agents} engineers in your team. Each engineer has a total of {max_rounds} rounds to complete their assigned tasks. Your responsibility is to delegate the reproduction tasks to these engineers and guide them to efficiently replicate as many of the core contributions and results of this paper as possible. Your team may not be able to complete everything within the allotted time; you should prioritize tasks in order to accomplish as much as possible within limited time. You may prepare the environment for the reproduction, but you are NOT allowed to generate any code for the reproduction. You may only delegate reproduction tasks to your engineers. Do NOT create code scaffolds, skeleton files, or markdown files -- your engineers will create their own implementation files.

Do not worry if your team fails to completely reproduce certain parts of the paper. Partial credit will be assigned. If a specific part of the reproduction is incorrect, only that part will fail to receive credit, while correctly implemented parts will be marked as successful. The partial credit will be weighted by importance to the paper's main contributions and results. Use your judgment to determine which parts of the paper are more important to reproduce. We will check that your results match the general trends of the original paper within a reasonable margin of error.

Experiments introduced only in the Appendix are not considered core contributions and are out of scope, even if referenced in the main text. However, if an experiment is described in the main body but some implementation details (e.g., prompts or long mathematical expressions) are placed in the Appendix, that experiment is still in scope.

RESOURCES
---
You may be provided with a rubric file located at `/workspace/paper/rubric.json`. The rubric has a hierarchical tree structure that breaks down the paper into fine-grained tasks. You may receive only a subset of the full hierarchy, or a single root-level task. In that case, use your judgment to decompose the reproduction into appropriate sub-tasks and assign them to your engineers. Do NOT delegate directly at rubric node level; the rubric is only a reference for expected outcomes.

An addendum file may also be available at `/workspace/paper/addendum.md`, providing additional context or clarifications regarding scope.

Your team may browse and use publicly available resources online. However, your team must NOT access the original paper's codebase or any resources listed in `blacklist.txt`.

SUBMISSION
---
Your team must produce a git repository located at `/workspace/submission/`. Only source code and necessary scripts should be committed; the total committed size must not exceed 1GB. We will run `git clean -fd` before grading, so any untracked files will be removed.

The submission MUST include a reproduction script at:
  /workspace/submission/reproduce.sh

We will copy your submission into a fresh Ubuntu 24.04 LTS Docker container and run:
  bash reproduce.sh

The maximum runtime is 7 days. Do not rely on hardcoded absolute paths. The container will include an NVIDIA A10 GPU with the NVIDIA container toolkit installed.

All graded artifacts must be generated by the reproduction script.

You should regularly update and test your reproduction script throughout development.

Finally, include a README.md file after all reproduction tasks are completed (or when engineers reach their maximum rounds). The README should summarize achieved results, explain the repository structure, and describe expected outputs from running reproduce.sh.

ENGINEER SUBMISSIONS
---
Each engineer must submit a local git commit upon completing their assigned task. If no commit is submitted, assume partial completion.

After reviewing a submission:
  1. Verify whether the expected outputs are produced.
  2. If incomplete, reassign the same task with corrective guidance.
  3. If complete, assign the next highest-priority remaining task.
  4. You may assign tasks to idle engineers to maximize productivity.
  5. Always summarize overall progress and provide detailed instructions for the next task.
  6. If onboarding a new engineer, provide a detailed explanation of the paper and current reproduction goals.

Output your response strictly in the following JSON format:

{
  "assign_task": {
    "reasoning": "Explain your decision",
    "tasks": [
      {
        "engineer_id": "engineer_id",
        "task_id": "task-unique-id",
        "task_node_id": "rubric task node id if available",
        "requirements": "Specific requirement to implement",
        "task_category": "Code Development|Experiment Running|Results Analysis|Other",
        "estimated_complexity": "simple|medium|complex",
        "instruction": "Provide detailed explanation of current progress and detailed instructions for this task, including expected behavior and outputs, relevant paper details, and required dependencies."
      }
    ]
  }
}

If no tasks should be assigned, use an empty tasks array.
\end{promptbox}

\begin{promptbox}{task delegation}
The engineers on your team are waiting for instructions to begin their first reproduction tasks. You must now delegate the reproduction work systematically by outputting a delegation JSON based on your current understanding of the paper. You have up to {max_agents} engineers available.

Strategies for effective first-round delegation:

- First, divide the overall reproduction effort into up to {max_agents} major task groups based on your understanding of the paper. Balance complexity and estimated effort as evenly as possible. Group related tasks together and carefully consider dependencies between tasks (i.e., which components depend on others). Do NOT delegate directly at the rubric node level; the rubric (if provided) is only a reference for expected outcomes. Remember that reproduction includes not only implementation but also experiment execution needed to generate expected outputs. When forming task groups, consider how experiment orchestration will be structured.

- For each engineer, assign the highest-priority reproduction task within their task group (i.e., the task that reproduces the most important results).

- Since this is the first assignment round, provide engineers with a clear explanation of the overall structure of the paper and a detailed summary of the paper based on your exploration. This ensures they do not need to re-explore the paper independently.

- Provide detailed instructions for each assigned task. Clearly specify which part of the paper is being reproduced and what outputs are expected. Include relevant context from the paper and addendum. Explicitly mention which dependencies are already available and which must be installed. Ensure that each engineer creates and modifies only their own files. Do NOT assign multiple engineers to modify the same file, as this will cause merge conflicts.

- Do not assign the reproduce.sh script to any engineer. You will create it yourself after all engineers have completed their tasks or reached their maximum rounds.

- Reproduction involves both implementation and experiment execution. Engineers must run experiments and generate concrete outputs (e.g., tables, figures, CSV files). Each task group should include both implementation and execution steps necessary to produce measurable results. The objective is to reproduce as many of the paper's core contributions and results as possible within limited time.

Note: Do NOT provide any code snippets or pseudo-code. Output your delegation plan strictly in the following JSON format:

{
  "delegation_plan": {
    "first_round": {
      "num_agents": <integer between 1 and {max_agents}>,
      "reasoning": "Explain why these tasks are prioritized and why this number of engineers is used.",
      "tasks": [
        {
          "engineer_id": "engineer_id",
          "task_id": "task-unique-id",
          "task_node_id": "rubric task node id if available",
          "requirements": "Specific requirement from the rubric to implement",
          "task_category": "Code Development|Experiment Running|Results Analysis|Other",
          "estimated_complexity": "simple|medium|complex",
          "instruction": "Provide a detailed explanation of the paper and detailed instructions for the current reproduction task. Explain the expected behavior and outputs. Include relevant details from the paper or addendum. Explicitly mention available dependencies and required installations. Provide clear, structured guidance to ensure correct implementation."
        }
      ]
    },
    "remaining_tasks": [
      {
        "task_id": "task-unique-id",
        "task_node_id": "rubric task node id if available",
        "requirements": "Specific requirement to implement",
        "task_category": "Code Development|Experiment Running|Results Analysis|Other",
        "estimated_complexity": "simple|medium|complex",
        "depends_on": ["list of task_ids this depends on, or empty"]
      }
    ]
  }
}
\end{promptbox}
\section{Full Results}
We provide the full results for each repository on Commit0-Lite and each paper on PaperBench across three LLMs.

\sisetup{
  detect-weight=true,
  detect-family=true,
  table-number-alignment=center,
  group-digits=false,
  table-format=3.1
}
\begin{table}[h]
\centering
\setlength{\tabcolsep}{4pt}
\renewcommand{\arraystretch}{1.08}
\begin{tabular}{
l
S[table-format=3.1] S[table-format=4.1] S[table-format=2.4] S[table-format=3.0]
S[table-format=3.1] S[table-format=4.1] S[table-format=2.4] S[table-format=3.0]
S[table-format=3.1] S[table-format=4.1] S[table-format=2.4] S[table-format=3.0]
}
\toprule
& \multicolumn{4}{c}{\textbf{Single-Agent (100 iters)}} 
& \multicolumn{4}{c}{\textbf{\ourmethod\ (4 engineers)}} 
& \multicolumn{4}{c}{\textbf{Single+\ourmethod}} \\
\cmidrule(lr){2-5} \cmidrule(lr){6-9} \cmidrule(lr){10-13}
\textbf{repo\_id}
& {\textbf{Pass}} & {\textbf{Time}} & {\textbf{Cost}} & {\textbf{Iter}}
& {\textbf{Pass}} & {\textbf{Time}} & {\textbf{Cost}} & {\textbf{Iter}}
& {\textbf{Pass}} & {\textbf{Time}} & {\textbf{Cost}} & {\textbf{Iter}} \\
\midrule
babel        & 0.0   & 955.3  & 1.2500  & 100 & 1.5   & 1749.4 & 13.9914 & 267 & 1.5   & 2704.7 & 15.2414 & 367 \\
cachetools   & 100.0 & 284.3  & 0.9926  & 44  & 100.0 & 863.1  & 3.3288  & 206 & 100.0 & 1147.4 & 4.3214  & 250 \\
chardet      & 6.4   & 598.0  & 2.2639  & 31  & 2.4   & 1112.4 & 7.4894  & 259 & 6.4   & 1710.4 & 9.7533  & 290 \\
cookiecutter & 35.1  & 615.8  & 1.8714  & 100 & 40.2  & 1246.9 & 6.9727  & 288 & 40.2  & 1862.7 & 8.8441  & 388 \\
deprecated   & 100.0 & 444.3  & 0.9812  & 47  & 100.0 & 1197.2 & 4.2408  & 165 & 100.0 & 1641.5 & 5.2220  & 212 \\
imapclient   & 28.8  & 596.6  & 1.9116  & 100 & 42.3  & 1463.0 & 9.0852  & 405 & 42.3  & 2059.6 & 10.9968 & 505 \\
jinja        & 0.0   & 647.2  & 1.7051  & 99  & 5.1   & 1483.9 & 9.9707  & 428 & 5.1   & 2131.1 & 11.6758 & 527 \\
marshmallow  & 23.1  & 600.6  & 2.0393  & 100 & 43.8  & 1981.0 & 10.9987 & 444 & 43.8  & 2581.6 & 13.0380 & 544 \\
minitorch    & 17.4  & 689.7  & 1.9825  & 100 & 34.4  & 1436.2 & 8.6874  & 374 & 34.4  & 2125.9 & 10.6699 & 474 \\
parsel       & 73.8  & 782.2  & 2.2379  & 97  & 72.3  & 1609.4 & 7.2702  & 275 & 73.8  & 2391.6 & 9.5081  & 372 \\
portalocker  & 79.0  & 1180.6 & 1.7070  & 78  & 100.0 & 2098.5 & 7.4321  & 275 & 100.0 & 3279.1 & 9.1391  & 353 \\
pyjwt        & 61.0  & 721.7  & 2.4165  & 99  & 62.2  & 1513.4 & 8.0366  & 330 & 62.2  & 2235.1 & 10.4531 & 429 \\
simpy        & 77.9  & 745.2  & 2.1196  & 100 & 92.1  & 2424.9 & 10.7432 & 387 & 92.1  & 3170.1 & 12.8628 & 487 \\
tinydb       & 91.0  & 838.6  & 2.5207  & 100 & 94.0  & 1730.0 & 7.1327  & 285 & 94.0  & 2568.6 & 9.6534  & 385 \\
voluptuous   & 56.4  & 747.5  & 2.5085  & 100 & 55.7  & 1801.3 & 9.2130  & 422 & 56.4  & 2548.8 & 11.7215 & 522 \\
wcwidth      & 100.0 & 634.4  & 1.6938  & 57  & 100.0 & 1620.2 & 5.5145  & 203 & 100.0 & 2254.6 & 7.2083  & 260 \\
\midrule
AVERAGE
& 53.1 & 692.6 & 1.8876 & 84.5
& 59.1 & 1583.2 & 8.1317 & 313.3
& 59.5 & 2275.8 & 10.0193 & 397.8 \\
\bottomrule
\end{tabular}
\caption{Claude 4.5 Sonnet results on Commit0-Lite across different configurations.}
\label{tab:claude-unified}
\end{table}

\begin{table}[h]
\centering
\setlength{\tabcolsep}{4pt}
\renewcommand{\arraystretch}{1.08}
\begin{tabular}{
l
S[table-format=3.1] S[table-format=4.1] S[table-format=2.1] S[table-format=3.0]
S[table-format=3.1] S[table-format=4.1] S[table-format=2.1] S[table-format=3.0]
S[table-format=3.1] S[table-format=4.1] S[table-format=2.1] S[table-format=3.0]
}
\toprule
& \multicolumn{4}{c}{\textbf{Single-Agent (100 iters)}} 
& \multicolumn{4}{c}{\textbf{Multi-Agent (4 engineers)}} 
& \multicolumn{4}{c}{\textbf{Single+Multi-Agent (100 iters)}} \\
\cmidrule(lr){2-5} \cmidrule(lr){6-9} \cmidrule(lr){10-13}
\textbf{repo\_id}
& {\textbf{Pass}} & {\textbf{Time}} & {\textbf{Cost}} & {\textbf{Iter}}
& {\textbf{Pass}} & {\textbf{Time}} & {\textbf{Cost}} & {\textbf{Iter}}
& {\textbf{Pass}} & {\textbf{Time}} & {\textbf{Cost}} & {\textbf{Iter}} \\
\midrule
babel        & 0.4   & 865.6  & 3.6  & 100 & 0.7   & 1658.4 & 11.8 & 395 & 0.7   & 2524.0 & 15.4 & 495 \\
cachetools   & 100.0 & 314.5  & 1.5  & 68  & 100.0 & 2131.2 & 4.1  & 179 & 100.0 & 2445.7 & 5.6  & 247 \\
chardet      & 0.0   & 438.0  & 4.1  & 100 & 0.0   & 1654.1 & 8.4  & 269 & 0.0   & 2092.1 & 12.5 & 369 \\
cookiecutter & 22.1  & 3817.7 & 3.2  & 100 & 29.0  & 2047.8 & 6.4  & 287 & 29.0  & 5865.5 & 9.6  & 387 \\
deprecated   & 100.0 & 210.1  & 0.9  & 44  & 100.0 & 2749.6 & 5.1  & 190 & 100.0 & 2959.7 & 6.0  & 234 \\
imapclient   & 23.2  & 550.4  & 2.6  & 100 & 24.3  & 1615.1 & 13.2 & 510 & 24.3  & 2165.5 & 15.7 & 610 \\
jinja        & 0.0   & 419.0  & 2.4  & 100 & 0.0   & 509.2  & 2.6  & 150 & 0.0   & 928.2  & 5.0  & 250 \\
marshmallow  & 17.0  & 392.6  & 2.3  & 100 & 38.7  & 2256.2 & 18.7 & 592 & 38.7  & 2648.8 & 21.0 & 692 \\
minitorch    & 17.4  & 555.5  & 2.3  & 100 & 20.0  & 744.7  & 9.2  & 372 & 20.0  & 1300.2 & 11.5 & 472 \\
parsel       & 39.8  & 486.3  & 2.6  & 100 & 47.6  & 552.8  & 5.3  & 240 & 47.6  & 1039.1 & 7.9  & 340 \\
portalocker  & 68.4  & 2957.4 & 1.3  & 56  & 71.1  & 1287.8 & 5.6  & 264 & 71.1  & 4245.2 & 6.8  & 320 \\
pyjwt        & 49.4  & 1039.5 & 3.1  & 100 & 59.5  & 527.4  & 1.5  & 54  & 59.5  & 1566.9 & 4.6  & 154 \\
simpy        & 34.3  & 672.4  & 2.6  & 100 & 65.0  & 1558.8 & 7.0  & 270 & 65.0  & 2231.2 & 9.6  & 370 \\
tinydb       & 82.1  & 458.9  & 3.2  & 100 & 71.6  & 1124.9 & 5.9  & 244 & 82.1  & 1583.8 & 9.1  & 344 \\
voluptuous   & 42.3  & 419.4  & 3.2  & 100 & 32.2  & 1052.4 & 6.2  & 246 & 42.3  & 1471.8 & 9.4  & 346 \\
wcwidth      & 89.5  & 338.9  & 0.8  & 30  & 84.2  & 734.2  & 5.5  & 181 & 89.5  & 1073.1 & 6.3  & 211 \\
\midrule
AVERAGE
& 42.9 & 871.0 & 2.5 & 87.4
& 46.5 & 1387.8 & 7.3 & 277.7
& 46.5 & 2258.8 & 9.8 & 365.1 \\
\bottomrule
\end{tabular}
\caption{GLM 4.7 results on Commit0-Lite across different configurations.}
\label{tab:glm47-commit0}
\end{table}

\begin{table}[h]
\centering

\setlength{\tabcolsep}{4pt}
\renewcommand{\arraystretch}{1.08}
\begin{tabular}{
l
S[table-format=3.1] S[table-format=4.1] S[table-format=1.1] S[table-format=3.0]
S[table-format=3.1] S[table-format=4.1] S[table-format=1.1] S[table-format=3.0]
S[table-format=3.1] S[table-format=4.1] S[table-format=1.1] S[table-format=3.0]
}
\toprule
& \multicolumn{4}{c}{\textbf{Single-Agent (100 iters)}} 
& \multicolumn{4}{c}{\textbf{Multi-Agent (4 engineers)}} 
& \multicolumn{4}{c}{\textbf{Single+Multi-Agent (100 iters)}} \\
\cmidrule(lr){2-5} \cmidrule(lr){6-9} \cmidrule(lr){10-13}
\textbf{repo\_id}
& {\textbf{Pass}} & {\textbf{Time}} & {\textbf{Cost}} & {\textbf{Iter}}
& {\textbf{Pass}} & {\textbf{Time}} & {\textbf{Cost}} & {\textbf{Iter}}
& {\textbf{Pass}} & {\textbf{Time}} & {\textbf{Cost}} & {\textbf{Iter}} \\
\midrule
babel        & 0.3   & 578.6  & 1.4 & 100 & 1.2   & 3972.7 & 9.4 & 514 & 1.2   & 4551.3 & 10.8 & 614 \\
cachetools   & 100.0 & 408.2  & 0.9 & 38  & 100.0 & 469.2  & 0.7 & 81  & 100.0 & 877.4  & 1.7  & 119 \\
chardet      & 3.5   & 612.3  & 1.7 & 64  & 31.7  & 2804.7 & 4.7 & 327 & 31.7  & 3417.0 & 6.3  & 391 \\
cookiecutter & 42.3  & 901.5  & 1.6 & 54  & 47.3  & 3593.6 & 6.8 & 407 & 47.3  & 4495.1 & 8.4  & 461 \\
deprecated   & 100.0 & 551.5  & 0.8 & 33  & 100.0 & 758.1  & 1.7 & 147 & 100.0 & 1309.6 & 2.5  & 180 \\
imapclient   & 18.0  & 443.9  & 1.1 & 100 & 16.9  & 871.1  & 1.5 & 31  & 18.0  & 1315.0 & 2.5  & 131 \\
jinja        & 0.0   & 419.5  & 3.6 & 100 & 0.0   & 1213.1 & 1.8 & 150 & 0.0   & 1632.6 & 5.3  & 250 \\
marshmallow  & 15.5  & 469.4  & 1.2 & 100 & 23.2  & 1217.8 & 5.4 & 242 & 23.2  & 1687.2 & 6.6  & 342 \\
minitorch    & 0.0   & 461.2  & 1.2 & 55  & 40.0  & 1164.6 & 2.0 & 112 & 40.0  & 1625.8 & 3.2  & 167 \\
parsel       & 52.9  & 857.6  & 1.8 & 52  & 100.0 & 1690.3 & 5.1 & 317 & 100.0 & 2547.9 & 6.9  & 369 \\
portalocker  & 76.3  & 978.6  & 1.7 & 70  & 97.4  & 3394.0 & 8.3 & 424 & 97.4  & 4372.6 & 10.0 & 494 \\
pyjwt        & 51.7  & 793.4  & 1.8 & 50  & 51.7  & 2385.2 & 7.9 & 424 & 51.7  & 3178.6 & 9.7  & 474 \\
simpy        & 0.0   & 1031.0 & 1.4 & 61  & 68.6  & 1578.1 & 5.6 & 138 & 68.6  & 2609.1 & 7.0  & 199 \\
tinydb       & 86.1  & 679.0  & 1.5 & 51  & 95.0  & 2817.1 & 6.1 & 171 & 95.0  & 3496.1 & 7.6  & 222 \\
voluptuous   & 37.6  & 919.5  & 1.5 & 69  & 38.3  & 1172.6 & 2.7 & 235 & 38.3  & 2092.1 & 4.2  & 304 \\
wcwidth      & 92.1  & 1927.9 & 2.8 & 39  & 100.0 & 1436.5 & 3.0 & 213 & 100.0 & 3364.4 & 5.8  & 252 \\
\midrule
AVERAGE
& 42.3 & 752.1 & 1.6 & 64.8
& 57.0 & 1908.7 & 4.5 & 245.8
& 57.0 & 2660.7 & 6.2 & 310.6 \\
\bottomrule
\end{tabular}
\caption{MiniMax 2.5 results on Commit0-Lite across different configurations.}
\label{tab:minimax25-commit0-heavy}
\end{table}
\sisetup{
  detect-weight=true,
  detect-family=true,
  group-digits=false,
  table-number-alignment=center
}
\begin{table*}[h]
\centering
\footnotesize
\setlength{\tabcolsep}{2.5pt}
\renewcommand{\arraystretch}{1.06}
\begin{tabular}{
>{\scriptsize\raggedright\arraybackslash}l
S[table-format=2.1]@{\hspace{3pt}}
S[table-format=4.1]@{\hspace{3pt}}
S[table-format=2.1]@{\hspace{4pt}}
S[table-format=3.1]@{\hspace{8pt}}
S[table-format=2.1]@{\hspace{3pt}}
S[table-format=4.1]@{\hspace{3pt}}
S[table-format=2.1]@{\hspace{4pt}}
S[table-format=3.1]@{\hspace{8pt}}
S[table-format=2.1]@{\hspace{3pt}}
S[table-format=4.1]@{\hspace{3pt}}
S[table-format=2.1]@{\hspace{4pt}}
S[table-format=3.1]
}
\toprule
& \multicolumn{4}{c}{\textbf{Single-Agent (100 iters)}}
& \multicolumn{4}{c}{\textbf{CAID (2 engineers)}}
& \multicolumn{4}{c}{\textbf{Single+CAID}} \\
\cmidrule(lr{6pt}){2-5} \cmidrule(lr{6pt}){6-9} \cmidrule(l){10-13}
\textbf{paper\_id}
& {\textbf{Scores}} & {\textbf{Time}} & {\textbf{Cost}} & {\textbf{Iter}}
& {\textbf{Scores}} & {\textbf{Time}} & {\textbf{Cost}} & {\textbf{Iter}}
& {\textbf{Scores}} & {\textbf{Time}} & {\textbf{Cost}} & {\textbf{Iter}} \\
\midrule
adaptive-pruning & 33.4 & 1043.5 & 3.0 & 70.0  & 56.0 & 2463.0 & 7.4 & 191.0 & 56.0 & 3506.5 & 10.5 & 261.0 \\
all-in-one & 68.4 & 3124.0 & 3.9 & 98.0  & 50.2 & 1946.9 & 6.0 & 146.0 & 68.4 & 5070.9 & 9.9  & 244.0 \\
bam & 57.9 & 3601.6 & 3.4 & 87.0  & 64.7 & 2577.7 & 7.2 & 223.0 & 64.7 & 6179.3 & 10.6 & 310.0 \\
bbox & 38.6 & 3397.5 & 4.0 & 80.0  & 68.8 & 1856.0 & 9.1 & 163.0 & 68.8 & 5253.5 & 13.1 & 243.0 \\
bridging-data-gaps & 43.2 & 1409.7 & 2.9 & 78.0  & 40.5 & 2078.0 & 6.6 & 166.0 & 43.2 & 3487.7 & 9.5  & 244.0 \\
fre & 56.9 & 1198.6 & 3.3 & 92.0  & 69.6 & 2193.6 & 7.6 & 213.0 & 69.6 & 3392.2 & 10.9 & 305.0 \\
ftrl & 34.6 & 1499.6 & 3.2 & 14.0  & 61.9 & 1943.0 & 7.3 & 184.0 & 61.9 & 3442.6 & 10.5 & 198.0 \\
lbcs & 79.5 & 1451.9 & 3.3 & 50.0  & 82.9 & 2508.9 & 6.3 & 170.0 & 82.9 & 3960.8 & 9.6  & 220.0 \\
lca-on-the-line & 59.3 & 1754.1 & 3.3 & 18.0  & 48.8 & 2011.9 & 4.7 & 205.0 & 59.3 & 3766.0 & 8.0  & 223.0 \\
mechanistic-understanding & 75.0 & 1771.8 & 3.0 & 77.0  & 63.1 & 1936.5 & 6.5 & 175.0 & 75.0 & 3708.3 & 9.5  & 252.0 \\
pinn & 53.9 & 2272.6 & 3.9 & 44.0  & 68.4 & 2222.5 & 5.3 & 112.0 & 68.4 & 4495.1 & 9.2  & 156.0 \\
rice & 33.2 & 2239.4 & 3.4 & 72.0  & 30.0 & 1870.7 & 6.4 & 150.0 & 33.2 & 4110.1 & 9.8  & 222.0 \\
robust-clip & 42.9 & 1343.8 & 3.4 & 83.0  & 57.2 & 1899.5 & 6.4 & 151.0 & 57.2 & 3243.3 & 9.7  & 234.0 \\
sample-specific-masks & 85.6 & 1110.3 & 2.7 & 22.0  & 86.3 & 2081.0 & 6.2 & 165.0 & 86.3 & 3191.3 & 8.9  & 187.0 \\
sapg & 28.0 & 1551.1 & 3.3 & 99.0  & 64.2 & 1934.8 & 8.0 & 139.0 & 64.2 & 3485.9 & 11.4 & 238.0 \\
sequential-neural-score-estimation & 86.5 & 2011.6 & 3.2 & 100.0 & 86.7 & 2097.5 & 4.7 & 164.0 & 86.7 & 4109.1 & 7.9  & 264.0 \\
stay-on-topic-with-classifier-free-guidance & 66.2 & 1468.2 & 3.0 & 62.0  & 78.5 & 1829.0 & 4.2 & 140.0 & 78.5 & 3297.2 & 7.2  & 202.0 \\
stochastic-interpolants & 85.8 & 1260.4 & 3.5 & 100.0 & 74.1 & 2105.3 & 6.6 & 217.0 & 85.8 & 3365.7 & 10.1 & 317.0 \\
test-time-model-adaptation & 62.7 & 1165.9 & 2.8 & 16.0  & 51.3 & 1966.1 & 6.2 & 165.0 & 62.7 & 3132.0 & 9.0  & 181.0 \\
what-will-my-model-forget & 52.4 & 1394.8 & 3.0 & 74.0  & 63.2 & 2086.1 & 6.5 & 126.0 & 63.2 & 3480.9 & 9.5  & 200.0 \\
\midrule
\bfseries AVERAGE
& 57.2 & 1803.5 & 3.3 & 66.8
& 63.3 & 2080.4 & 6.5 & 168.3
& 66.8 & 3883.9 & 9.7 & 235.1 \\
\bottomrule
\end{tabular}
\caption{Claude 4.5 Sonnet results on PaperBench Code-Dev across different configurations.}
\label{tab:claude45-paperbench-code-dev}
\end{table*}

\begin{table*}[h]
\centering
\footnotesize
\setlength{\tabcolsep}{2.5pt}
\renewcommand{\arraystretch}{1.06}
\begin{tabular}{
>{\scriptsize\raggedright\arraybackslash}l
S[table-format=2.1]@{\hspace{3pt}}
S[table-format=4.1]@{\hspace{3pt}}
S[table-format=2.1]@{\hspace{4pt}}
S[table-format=3.1]@{\hspace{8pt}}
S[table-format=2.1]@{\hspace{3pt}}
S[table-format=4.1]@{\hspace{3pt}}
S[table-format=2.1]@{\hspace{4pt}}
S[table-format=3.1]@{\hspace{8pt}}
S[table-format=2.1]@{\hspace{3pt}}
S[table-format=4.1]@{\hspace{3pt}}
S[table-format=2.1]@{\hspace{4pt}}
S[table-format=3.1]
}
\toprule
& \multicolumn{4}{c}{\textbf{Single-Agent (100 iters)}}
& \multicolumn{4}{c}{\textbf{Multi-Agent (2 engineers)}}
& \multicolumn{4}{c}{\textbf{Single+Multi-Agent}} \\
\cmidrule(lr{6pt}){2-5} \cmidrule(lr{6pt}){6-9} \cmidrule(l){10-13}
\textbf{paper\_id}
& {\textbf{Scores}} & {\textbf{Time}} & {\textbf{Cost}} & {\textbf{Iter}}
& {\textbf{Scores}} & {\textbf{Time}} & {\textbf{Cost}} & {\textbf{Iter}}
& {\textbf{Scores}} & {\textbf{Time}} & {\textbf{Cost}} & {\textbf{Iter}} \\
\midrule
adaptive-pruning & 44.9 & 1130.0 & 2.5 & 72.0  & 60.3 & 1473.6 & 6.1 & 187.0 & 60.3 & 2603.6 & 8.6 & 259.0 \\
all-in-one & 19.9 & 1430.0 & 3.6 & 100.0 & 25.8 & 1532.0 & 3.3 & 140.0 & 25.8 & 2962.0 & 6.9 & 240.0 \\
bam & 63.5 & 681.6 & 2.5 & 53.0  & 75.3 & 1315.1 & 4.9 & 184.0 & 75.3 & 1996.7 & 7.4 & 237.0 \\
bbox & 15.1 & 1186.0 & 2.7 & 75.0  & 40.1 & 1227.9 & 4.4 & 163.0 & 40.1 & 2413.9 & 7.1 & 238.0 \\
bridging-data-gaps & 25.6 & 603.0 & 2.1 & 68.0  & 33.5 & 1227.1 & 4.8 & 190.0 & 33.5 & 1830.1 & 6.9 & 258.0 \\
fre & 42.3 & 2429.6 & 2.6 & 55.0  & 42.8 & 1349.6 & 4.4 & 177.0 & 42.8 & 3779.2 & 7.0 & 232.0 \\
ftrl & 15.4 & 1326.4 & 3.1 & 95.0  & 32.7 & 1850.0 & 5.2 & 182.0 & 32.7 & 3176.4 & 8.3 & 277.0 \\
lbcs & 75.0 & 539.7 & 3.3 & 87.0  & 38.2 & 1213.2 & 4.5 & 145.0 & 75.0 & 1752.9 & 7.8 & 232.0 \\
lca-on-the-line & 34.7 & 675.7 & 2.9 & 58.0  & 30.2 & 1974.9 & 3.2 & 112.0 & 34.7 & 2650.6 & 6.1 & 170.0 \\
mechanistic-understanding & 0.0 & 3601.7 & 3.3 & 90.0  & 47.7 & 1904.2 & 3.6 & 164.0 & 47.7 & 5505.9 & 6.9 & 254.0 \\
pinn & 61.0 & 1158.6 & 2.4 & 43.0  & 43.2 & 832.6 & 4.3 & 155.0 & 61.0 & 1991.2 & 6.7 & 198.0 \\
rice & 28.5 & 867.8 & 3.4 & 96.0  & 30.0 & 1870.7 & 3.6 & 131.0 & 30.0 & 2738.5 & 7.0 & 227.0 \\
robust-clip & 22.3 & 728.7 & 3.7 & 87.0  & 29.3 & 1288.9 & 7.1 & 191.0 & 29.3 & 2017.6 & 10.9 & 278.0 \\
sample-specific-masks & 50.4 & 793.3 & 2.4 & 58.0  & 54.6 & 1123.5 & 5.5 & 217.0 & 54.6 & 1916.8 & 7.9 & 275.0 \\
sapg & 29.4 & 836.0 & 4.5 & 100.0 & 27.0 & 952.0 & 6.0 & 204.0 & 29.4 & 1788.0 & 10.5 & 304.0 \\
sequential-neural-score-estimation & 58.8 & 1248.6 & 2.8 & 92.0  & 79.9 & 1136.1 & 4.4 & 176.0 & 79.9 & 2384.7 & 7.2 & 268.0 \\
stay-on-topic-with-classifier-free-guidance & 49.7 & 807.1 & 2.6 & 81.0  & 59.3 & 1769.4 & 4.8 & 157.0 & 59.3 & 2576.5 & 7.4 & 238.0 \\
stochastic-interpolants & 70.8 & 1376.5 & 3.0 & 67.0  & 71.0 & 1586.8 & 6.7 & 228.0 & 71.0 & 2963.3 & 9.7 & 295.0 \\
test-time-model-adaptation & 10.3 & 1106.9 & 1.0 & 92.0  & 32.9 & 1547.6 & 3.2 & 133.0 & 32.9 & 2654.5 & 4.2 & 225.0 \\
what-will-my-model-forget & 42.6 & 1023.9 & 1.9 & 61.0  & 53.6 & 1812.9 & 4.6 & 76.0  & 53.6 & 2836.8 & 6.4 & 137.0 \\
\midrule
\bfseries AVERAGE
& 38.0 & 1177.6 & 2.8 & 76.5
& 45.4 & 1449.4 & 4.7 & 165.6
& 48.5 & 2627.0 & 7.5 & 242.3 \\
\bottomrule
\end{tabular}
\caption{GLM 4.7 results on PaperBench Code-Dev across different configurations.}
\label{tab:glm47-paperbench}
\end{table*}

\begin{table*}[h]
\centering
\footnotesize
\setlength{\tabcolsep}{2.5pt}
\renewcommand{\arraystretch}{1.06}
\begin{tabular}{
>{\scriptsize\raggedright\arraybackslash}l
S[table-format=2.1]@{\hspace{3pt}}
S[table-format=4.1]@{\hspace{3pt}}
S[table-format=1.1]@{\hspace{4pt}}
S[table-format=3.1]@{\hspace{8pt}}
S[table-format=2.1]@{\hspace{3pt}}
S[table-format=4.1]@{\hspace{3pt}}
S[table-format=1.1]@{\hspace{4pt}}
S[table-format=3.1]@{\hspace{8pt}}
S[table-format=2.1]@{\hspace{3pt}}
S[table-format=4.1]@{\hspace{3pt}}
S[table-format=1.1]@{\hspace{4pt}}
S[table-format=3.1]
}
\toprule
& \multicolumn{4}{c}{\textbf{Single-Agent (100 iters)}}
& \multicolumn{4}{c}{\textbf{Multi-Agent (2 engineers)}}
& \multicolumn{4}{c}{\textbf{Single+Multi-Agent}} \\
\cmidrule(lr{6pt}){2-5} \cmidrule(lr{6pt}){6-9} \cmidrule(l){10-13}
\textbf{paper\_id}
& {\textbf{Scores}} & {\textbf{Time}} & {\textbf{Cost}} & {\textbf{Iter}}
& {\textbf{Scores}} & {\textbf{Time}} & {\textbf{Cost}} & {\textbf{Iter}}
& {\textbf{Scores}} & {\textbf{Time}} & {\textbf{Cost}} & {\textbf{Iter}} \\
\midrule
adaptive-pruning & 15.1 & 3601.5 & 0.9 & 50.0  & 15.2 & 3558.3 & 3.1 & 223.0 & 15.2 & 7159.8 & 4.0 & 273.0 \\
all-in-one & 0.0 & 2461.6 & 1.2 & 35.0  & 22.3 & 3635.1 & 2.7 & 198.0 & 22.3 & 6096.7 & 3.9 & 233.0 \\
bam & 49.9 & 2434.8 & 1.3 & 11.0  & 38.5 & 1852.4 & 2.6 & 216.0 & 49.9 & 4287.2 & 3.8 & 227.0 \\
bbox & 0.0 & 1491.0 & 0.5 & 41.0  & 28.0 & 1257.8 & 1.3 & 136.0 & 28.0 & 2748.8 & 1.9 & 177.0 \\
bridging-data-gaps & 29.4 & 970.1 & 0.8 & 57.0  & 33.5 & 2610.4 & 2.7 & 181.0 & 33.5 & 3580.5 & 3.5 & 238.0 \\
fre & 0.0 & 2128.3 & 2.5 & 100.0 & 29.0 & 2955.0 & 4.3 & 284.0 & 29.0 & 5083.3 & 6.8 & 384.0 \\
ftrl & 0.0 & 3601.2 & 0.9 & 55.0  & 7.1 & 4130.4 & 2.4 & 193.0 & 7.1 & 7731.6 & 3.4 & 248.0 \\
lbcs & 0.0 & 3601.1 & 0.6 & 36.0  & 0.4 & 2933.2 & 3.0 & 189.0 & 42.0 & 6534.3 & 3.6 & 225.0 \\
lca-on-the-line & 11.8 & 1045.4 & 0.6 & 45.0  & 0.2 & 3294.7 & 3.3 & 239.0 & 24.9 & 4340.1 & 3.9 & 284.0 \\
mechanistic-understanding & 0.0 & 3600.7 & 1.1 & 64.0  & 0.3 & 3129.2 & 1.3 & 131.0 & 34.0 & 6729.9 & 2.4 & 195.0 \\
pinn & 0.0 & 1906.9 & 1.2 & 67.0  & 0.6 & 2714.2 & 2.6 & 192.0 & 56.0 & 4621.1 & 3.8 & 259.0 \\
rice & 0.0 & 3601.6 & 0.8 & 53.0  & 0.2 & 2509.7 & 2.2 & 173.0 & 20.6 & 6111.3 & 3.0 & 226.0 \\
robust-clip & 0.0 & 2474.9 & 1.4 & 75.0  & 0.2 & 3668.2 & 3.2 & 250.0 & 23.9 & 6143.1 & 4.6 & 325.0 \\
sample-specific-masks & 0.0 & 3601.1 & 0.6 & 46.0  & 0.6 & 4419.1 & 1.8 & 78.0  & 58.7 & 8020.2 & 2.4 & 124.0 \\
sapg & 8.4 & 1780.2 & 0.6 & 46.0  & 0.3 & 1934.8 & 0.9 & 150.0 & 29.9 & 3715.0 & 1.5 & 196.0 \\
sequential-neural-score-estimation & 47.4 & 2511.2 & 1.0 & 75.0  & 0.7 & 3759.2 & 3.0 & 137.0 & 71.1 & 6270.4 & 4.0 & 212.0 \\
stay-on-topic-with-classifier-free-guidance & 0.5 & 2882.9 & 1.2 & 71.0  & 0.5 & 3029.0 & 3.2 & 176.0 & 0.5 & 5911.9 & 4.5 & 247.0 \\
stochastic-interpolants & 0.0 & 2426.3 & 2.1 & 100.0 & 0.7 & 3608.0 & 4.8 & 279.0 & 0.7 & 6034.3 & 6.8 & 379.0 \\
test-time-model-adaptation & 0.0 & 2990.1 & 1.5 & 93.0  & 0.5 & 1989.6 & 1.6 & 137.0 & 0.5 & 4979.7 & 3.1 & 230.0 \\
what-will-my-model-forget & 0.0 & 1395.7 & 0.9 & 45.0  & 0.2 & 3859.7 & 2.0 & 49.0  & 0.2 & 5255.4 & 2.9 & 94.0 \\
\midrule
\bfseries AVERAGE
& 10.5 & 2525.3 & 1.1 & 58.3
& 36.1 & 3042.4 & 2.6 & 180.6
& 36.7 & 5567.7 & 3.7 & 238.8 \\
\bottomrule
\end{tabular}
\caption{MiniMax 2.5 results on PaperBench Code-Dev across different configurations.}
\label{tab:minimax25-paperbench}
\end{table*}

\section{One-sided t-test}
\label{app:stats-test}
\begin{wraptable}{r}{0.5\textwidth}
\vspace{-12pt}
\centering
\small
\begin{tabular}{llccc}
\toprule
Benchmark & Model & $\Delta$ & $t$ & $p$ \\
\midrule
\multirow{3}{*}{Commit0}
 & Claude 4.5  & +6.0 & 2.87 & 0.006 \\
 & GLM 4.7     & +3.6 & 1.37 & 0.095 \\
 & MiniMax 2.5 & +14.7 & 2.81 & 0.007 \\
\midrule
\multirow{3}{*}{PaperBench}
 & Claude 4.5  & +6.1 & 1.78 & 0.046 \\
 & GLM 4.7     & +7.4 & 1.93 & 0.034 \\
 & MiniMax 2.5 & +25.6 & 5.27 & <0.0001 \\
\bottomrule
\end{tabular}
\caption{One-sided paired $t$-test ($H_1$: \ourmethod $>$ Single-Agent). $\Delta$: mean score improvement. Bold: $p < 0.05$.}
\label{tab:significance}
\vspace{-10pt}
\end{wraptable}
 
We compute one-sided paired $t$-tests ($H_1$: \ourmethod $>$ Single-Agent) across all repositories or papers for each model in Table~\ref{tab:significance}. On Commit0-Lite, the improvement is significant for Claude Sonnet 4.5 ($t=2.87$, $p=0.006$) and MiniMax 2.5 ($t=2.81$, $p=0.007$), with mean gains of 6.0 and 14.7 percentage points respectively. GLM 4.7 improves by 3.6 points on average but does not reach significance ($p=0.095$), largely because the per-repository variance is high: \ourmethod brings large gains on some repositories (e.g., +30.7 on \texttt{simpy}) but regresses on others (e.g., $-10.5$ on \texttt{tinydb}), which inflates the standard error with only 16 paired samples. On PaperBench, all three models, Claude Sonnet 4.5 ($t=1.78$, $p=0.046$), GLM 4.7 ($t=1.93$, $p=0.034$) and MiniMax 2.5 ($t=5.27$, $p<0.0001$) are significant. 
As discussed in Section~\ref{sec:delegation}, \ourmethod's effectiveness depends on the manager's ability to construct accurate dependency graphs and delegate tasks accordingly. A weaker base model produces less reliable task decomposition on the open-ended PaperBench tasks, limiting the gains that multi-agent coordination can deliver. 
\section{Failure on Scaling the Parallel Execution}
\label{app:failure-parallel}
\begin{figure*}[h!]
    \centering
    \small
    \includegraphics[width=0.9\textwidth]{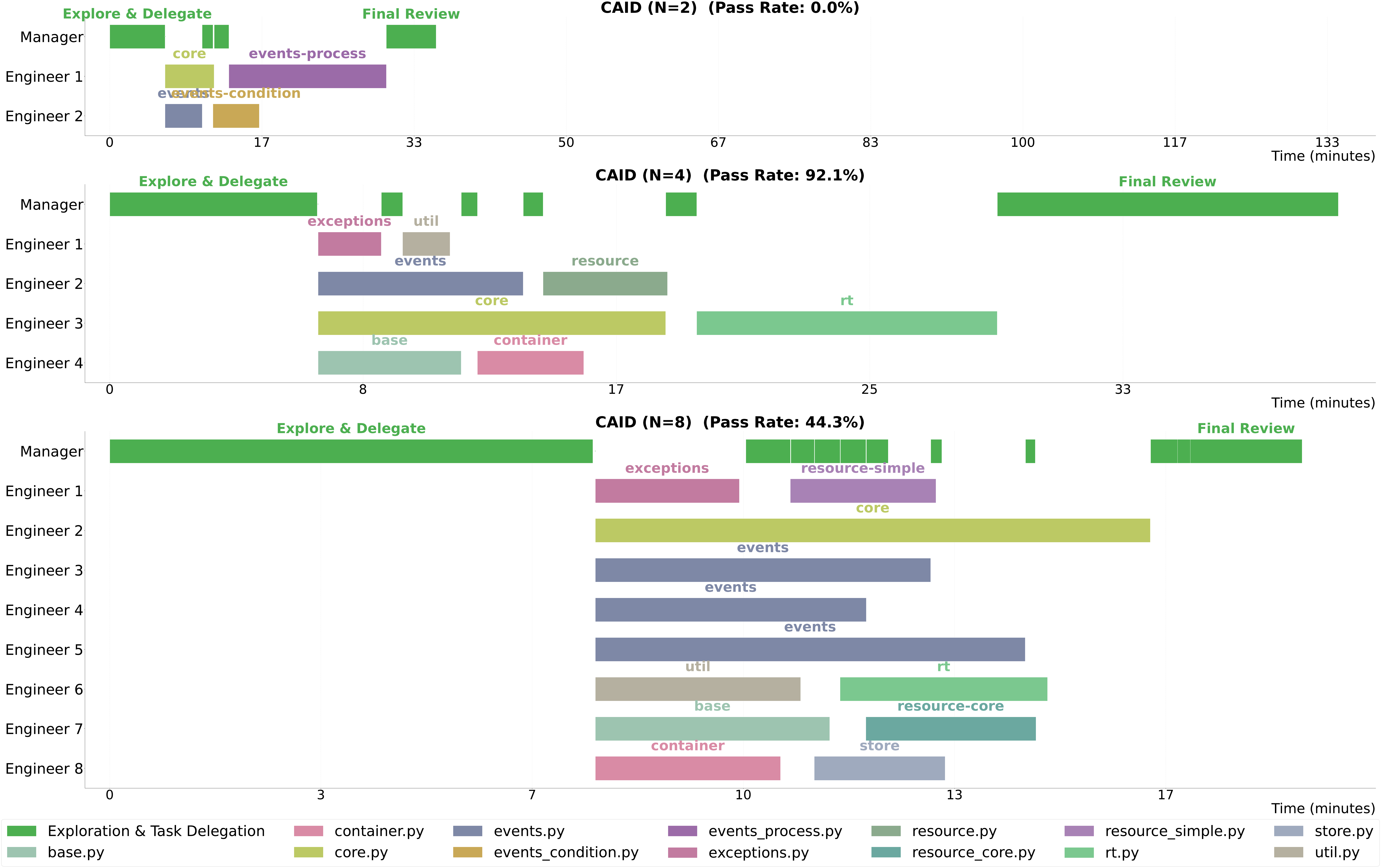}
    \caption{Gantt plot on the \texttt{simpy} repository for \ourmethod with different number of engineers, where $N=2,4,8$. }
    \label{fig:parallel-execution-simpy}
\end{figure*}

We provide an example to show why scaling parallel execution does not always help. Figure~\ref{fig:parallel-execution-simpy} shows the execution timelines on the \texttt{simpy} repository under different numbers of engineers ($N=2,4,8$). The performance difference is not solely explained by the number of files touched, but by how the manager structures delegation across engineers. For $N=4$, delegation remains clean and non-overlapping. Each engineer is assigned distinct files (e.g., \texttt{events.py}, \texttt{core.py}, \texttt{container.py}, \texttt{resource.py}), and their implementations proceed largely without interference. The manager avoids assigning closely coupled modules to different engineers simultaneously, and no two engineers work on the same file at the same time. As a result, integration remains stable and the run reaches a pass rate of 92.1\%.

For $N=8$, although more files are modified and parallel activity increases, the delegation becomes less disciplined. Multiple engineers are assigned different functions within the same file (notably \texttt{events.py}), creating overlapping write regions within a shared module. While these edits are logically separable at the function level, they introduce integration risk at the file level. The main branch receives competing updates on the same module, increasing the likelihood of merge conflicts or inconsistent intermediate states. This fragmentation of responsibility prevents clean consolidation and ultimately limits performance to 44.3\%. The degradation in $N=8$ therefore does not arise from excessive parallelism alone, but from a delegation that ignores the ownership boundaries of the file-level. When parallel execution exceeds the manager’s ability to enforce coherent task partitioning, local productivity no longer translates into stable global progress. This example illustrates that scaling the number of engineers requires disciplined delegation, not simply increasing concurrency.

\end{document}